\ifdefined\pdftexversion
\pdfoutput=1
\fi
\documentclass{article}

\PassOptionsToPackage{square}{natbib}
\usepackage[preprint]{neurips_2026}

\usepackage[utf8]{inputenc}
\usepackage[T1]{fontenc}

\usepackage{amsmath}
\usepackage{newtxmath}
\usepackage{float}
\usepackage{hyperref}
\usepackage[figure]{hypcap}
\usepackage{url}
\mathchardef\UrlBigBreakPenalty=9000
\usepackage{booktabs}
\usepackage[expansion=false]{microtype}
\usepackage{xcolor}
\usepackage{graphicx}
\usepackage{tabularx}
\usepackage{array}
\usepackage{threeparttable}
\usepackage{enumitem}
\ifdefined\pdfsuppressptexinfo\pdfsuppressptexinfo=-1\fi
\definecolor{linkblue}{HTML}{1C5CAB}
\hypersetup{
  colorlinks=true, linkcolor=linkblue, citecolor=linkblue, urlcolor=linkblue,
  bookmarksnumbered=true, pdflang={en-US},
  pdftitle={What RL on Expert-Built Tasks Teaches Nemotron 3.5 Super},
  pdfauthor={Sushant Mehta, Edwin Chen},
  pdfsubject={Reinforcement learning on expert-built task datasets},
  pdfkeywords={reinforcement learning, LoRA, CISPO, Nemotron, agents, instruction following, charts, documents}
}

\newcommand{\ra}{\ensuremath{\,\rightarrow\,}}
\newcolumntype{Y}{>{\raggedright\arraybackslash}X}

\newcommand{\ModelTotalParams}{120}
\newcommand{\ModelActiveParams}{12}
\newcommand{\CanaryMeanDiff}{0.40}
\newcommand{\CanaryTokens}{8,319}
\newcommand{\DeployIdEarly}{nvidia/NVIDIA-Nemotron-3.5-Super-VL-120B-A12B-BF16:peft:262144:ea}
\newcommand{\DeployIdPlain}{nvidia/NVIDIA-Nemotron-3.5-Super-VL-120B-A12B-BF16:peft:262144}
\newcommand{\RecipeLoRARankSWE}{32}
\newcommand{\RecipeClipHigh}{4}
\newcommand{\RecipeClipLow}{0}
\newcommand{\RecipeAdamBetaOne}{0.9}
\newcommand{\RecipeAdamBetaTwo}{0.95}
\newcommand{\RecipeGradClip}{1.0}
\newcommand{\RecipeWeightDecay}{0}
\newcommand{\RecipeAdamEps}{\ensuremath{1\times10^{-8}}}
\newcommand{\RecipeLRSWE}{\ensuremath{2\times10^{-4}}}
\newcommand{\RecipeLRConstraints}{\ensuremath{1\times10^{-4}}}
\newcommand{\RecipeLRToolUse}{\ensuremath{1\times10^{-4}}}
\newcommand{\RecipeLRCharts}{\ensuremath{1\times10^{-4}}}
\newcommand{\RecipeLRDocs}{\ensuremath{1\times10^{-4}}}
\newcommand{\RecipeStalenessSWE}{4}
\newcommand{\RecipeStalenessConstraints}{3}
\newcommand{\RecipeStalenessToolUse}{2}
\newcommand{\RecipeStalenessCharts}{8}
\newcommand{\RecipeStalenessDocs}{2}
\newcommand{\RecipeMaxTokensConstraints}{49,152}
\newcommand{\RecipeTurnTokensToolUse}{32,768}
\newcommand{\RecipeTurnTokensCharts}{32,768}
\newcommand{\RecipeContextCharts}{258,048}
\newcommand{\RecipeTurnCapCharts}{50}
\newcommand{\RecipeTurnTokensDocs}{32,768}
\newcommand{\RecipeTurnCapDocs}{20}
\newcommand{\RecipeTurnCapSWE}{250}
\newcommand{\RecipeEpisodeHoursSWE}{5}
\newcommand{\RecipeEpisodeHoursToolUse}{3}
\newcommand{\RecipeTrainTemperature}{1.0}
\newcommand{\RecipeTrainTopP}{1.0}
\newcommand{\RecipeOverlongAmpConstraints}{0.25}
\newcommand{\RecipeOverlongStartConstraints}{32,768}
\newcommand{\RecipeOverlongWidthConstraints}{16,384}
\newcommand{\RecipeRewardAllPassWeightConstraints}{0.5}
\newcommand{\RecipeRewardDocs}{0.5}
\newcommand{\RecipeWrapupFloorToolUse}{0.2}
\newcommand{\RecipeRunoutChargeToolUse}{0.1}
\newcommand{\RecipeRunoutStartToolUse}{200,000}
\newcommand{\RecipeLeversFractionToolUse}{0.3}
\newcommand{\RecipeTurnTokensSWE}{32,768}
\newcommand{\RecipeContextSWE}{258,048}
\newcommand{\RecipeContextToolUse}{258,048}
\newcommand{\RecipeContextDocs}{258,048}
\newcommand{\RecipeTurnCapToolUse}{200}
\newcommand{\RecipeViolationScaleToolUse}{0.25}
\newcommand{\RecipeRewardMinToolUse}{\ensuremath{-0.25}}
\newcommand{\RecipeLongTurnChargeToolUse}{0.1}
\newcommand{\RecipeLongTurnStartToolUse}{6,144}
\newcommand{\RecipeLongTurnWidthToolUse}{10,240}
\newcommand{\RecipeCutoffFloorToolUse}{0.1}
\newcommand{\RecipeCutoffFloorDocs}{0.1}
\newcommand{\RecipeForcedScaleToolUse}{0.5}
\newcommand{\CMNoOverlapGain}{\ensuremath{+6.46}}
\newcommand{\CMNoOverlapCI}{\ensuremath{[+4.86, +8.12]}}
\newcommand{\SizeSWEPro}{635}
\newcommand{\SizeSWEProTotal}{642}
\newcommand{\SizeTBench}{89}
\newcommand{\SizeDeepSWE}{112}
\newcommand{\SizeDeepSWETotal}{113}
\newcommand{\SizeCC}{75}
\newcommand{\SizeAIF}{1,643}
\newcommand{\SizeAIFTotal}{1,645}
\newcommand{\SizeIFB}{300}
\newcommand{\SizeMC}{266}
\newcommand{\SizeAB}{600}
\newcommand{\SizeHB}{65}
\newcommand{\SizeChart}{100}
\newcommand{\SizeCM}{1,000}
\newcommand{\SizeGDP}{100}
\newcommand{\SizeSWEVerifiedTotal}{500}
\newcommand{\SWEProAttempts}{2}
\newcommand{\TBenchAttempts}{5}
\newcommand{\DeepSWEAttempts}{4}
\newcommand{\IFBAttempts}{8}
\newcommand{\MCAttempts}{8}
\newcommand{\AIFAttempts}{2}
\newcommand{\ABAttempts}{1}
\newcommand{\HBAttempts}{5}
\newcommand{\ChartAttempts}{10}
\newcommand{\CMAttempts}{2}
\newcommand{\CCAttempts}{8}
\newcommand{\GDPAttempts}{5}
\newcommand{\EvalTemperature}{1.0}
\newcommand{\EvalTopP}{0.95}
\newcommand{\ABTopP}{1.0}
\newcommand{\SWECallCap}{250}
\newcommand{\SWETurnTokens}{65,536}
\newcommand{\TBenchToolHours}{3}
\newcommand{\TBenchBackstopHours}{10}
\newcommand{\DeepSWECallCap}{1,000}
\newcommand{\ABTurnLimit}{50}
\newcommand{\LongHorizonTurnTokens}{32,768}
\newcommand{\HBTurnCap}{1,000}
\newcommand{\HBAgentMinutes}{120}
\newcommand{\ChartTurnCap}{50}
\newcommand{\ChartContextTokens}{258,048}
\newcommand{\InstructionOutputCap}{131,072}
\newcommand{\GDPEvalTurnCap}{40}
\newcommand{\SWEVerifiedDropped}{15}
\newcommand{\SWEVerifiedDroppedInternet}{9}
\newcommand{\SWEVerifiedDroppedEmptyPatch}{2}
\newcommand{\SWEVerifiedDroppedGolden}{3}
\newcommand{\SWEVerifiedDroppedInfra}{1}
\newcommand{\BootResamples}{10,000}
\newcommand{\GDPMaskedCriteria}{49}
\newcommand{\GDPRejudgedCriteria}{27}
\newcommand{\CCDevGainSecond}{\ensuremath{+11.6}}
\newcommand{\CCDevGainSecondCI}{\ensuremath{[+8.1, +15.3]}}
\newcommand{\OvlTBenchAnyThirteen}{0}
\newcommand{\OvlTBenchMaxShare}{0.0}
\newcommand{\OvlSWEProAnyThirteen}{0}
\newcommand{\OvlSWEProMaxShare}{1.0}
\newcommand{\OvlDeepSWEAnyThirteen}{0}
\newcommand{\OvlDeepSWEMaxShare}{0.4}
\newcommand{\OvlSWEVerifiedAnyThirteen}{0}
\newcommand{\OvlSWEVerifiedMaxShare}{6.6}
\newcommand{\OvlIFBAnyThirteen}{0}
\newcommand{\OvlIFBMaxShare}{0.0}
\newcommand{\OvlCCAnyThirteen}{2}
\newcommand{\OvlCCMaxShare}{28.4}
\newcommand{\OvlAIFAnyThirteen}{1}
\newcommand{\OvlAIFMaxShare}{5.1}
\newcommand{\OvlMCAnyThirteen}{0}
\newcommand{\OvlMCMaxShare}{2.1}
\newcommand{\OvlABAnyThirteen}{0}
\newcommand{\OvlABMaxShare}{0.9}
\newcommand{\OvlCMAnyThirteen}{0}
\newcommand{\OvlCMMaxShare}{28.6}
\newcommand{\OvlChartAnyThirteen}{0}
\newcommand{\OvlChartMaxShare}{3.7}
\newcommand{\OvlCMWithoutFlaggedGain}{\ensuremath{+6.51}}
\newcommand{\OvlCMWithoutFlaggedCI}{\ensuremath{[+4.85, +8.16]}}
\newcommand{\OvlCCTwins}{3}
\newcommand{\OvlCCWithoutThirteenGain}{\ensuremath{+7.6}}
\newcommand{\OvlCCWithoutThirteenCI}{\ensuremath{[+1.6, +13.7]}}
\newcommand{\OvlAIFWithoutThirteenGain}{\ensuremath{+7.8}}
\newcommand{\OvlAIFWithoutThirteenCI}{\ensuremath{[+5.8, +9.7]}}
\newcommand{\OvlHBSeedsMaxShare}{4.6}
\newcommand{\OvlHBWithoutThirteenGain}{\ensuremath{+11.0}}
\newcommand{\OvlHBWithoutThirteenCI}{\ensuremath{[+7.4, +14.9]}}
\newcommand{\OvlHBThirteenTasks}{27}
\newcommand{\OvlGDPCriteriaMaxShare}{5.8}
\newcommand{\OvlPlantedCopies}{28}
\newcommand{\OvlPlantedLongFlagged}{22}
\newcommand{\OvlCMSharedCharts}{2}
\newcommand{\SWEProBase}{86.9}
\newcommand{\SWEProTrained}{92.1}
\newcommand{\SWEProGain}{\ensuremath{+5.2}}
\newcommand{\SWEProCI}{\ensuremath{[+3.3, +7.1]}}
\newcommand{\TBenchBase}{63.9}
\newcommand{\TBenchTrained}{70.3}
\newcommand{\TBenchGain}{\ensuremath{+6.3}}
\newcommand{\TBenchCI}{\ensuremath{[+2.4, +10.6]}}
\newcommand{\DeepSWEBase}{27.9}
\newcommand{\DeepSWETrained}{32.6}
\newcommand{\DeepSWEGain}{\ensuremath{+4.7}}
\newcommand{\DeepSWECI}{\ensuremath{[-0.7, +9.8]}}
\newcommand{\CCBase}{27.3}
\newcommand{\CCTrained}{35.5}
\newcommand{\CCGain}{\ensuremath{+8.2}}
\newcommand{\CCCI}{\ensuremath{[+2.3, +14.0]}}
\newcommand{\CCNoTwinsGain}{\ensuremath{+7.1}}
\newcommand{\CCNoTwinsCI}{\ensuremath{[+1.0, +13.2]}}
\newcommand{\AIFBase}{50.6}
\newcommand{\AIFTrained}{58.3}
\newcommand{\AIFGain}{\ensuremath{+7.7}}
\newcommand{\AIFCI}{\ensuremath{[+5.8, +9.7]}}
\newcommand{\IFBBase}{72.3}
\newcommand{\IFBTrained}{76.8}
\newcommand{\IFBGain}{\ensuremath{+4.5}}
\newcommand{\IFBCI}{\ensuremath{[+2.4, +6.7]}}
\newcommand{\IFBPromptLooseGain}{\ensuremath{+4.5}}
\newcommand{\IFBPromptLooseCI}{\ensuremath{[+2.4, +6.7]}}
\newcommand{\MCBase}{58.3}
\newcommand{\MCTrained}{66.6}
\newcommand{\MCGain}{\ensuremath{+8.3}}
\newcommand{\MCCI}{\ensuremath{[+5.5, +11.2]}}
\newcommand{\ABBase}{14.8}
\newcommand{\ABTrained}{27.5}
\newcommand{\ABGain}{\ensuremath{+12.7}}
\newcommand{\ABCI}{\ensuremath{[+9.3, +16.2]}}
\newcommand{\ABOutcomeBase}{31.9}
\newcommand{\ABOutcomeTrained}{49.5}
\newcommand{\ABOutcomeGain}{\ensuremath{+17.6}}
\newcommand{\ABOutcomeCI}{\ensuremath{[+14.3, +20.9]}}
\newcommand{\ABChecksGain}{\ensuremath{+20.6}}
\newcommand{\ABChecksCI}{\ensuremath{[+17.7, +23.6]}}
\newcommand{\ABGuardBase}{37.3}
\newcommand{\ABGuardTrained}{32.5}
\newcommand{\ABGuardGain}{\ensuremath{-4.8}}
\newcommand{\ABGuardCI}{\ensuremath{[-8.5, -1.2]}}
\newcommand{\HBBase}{54.8}
\newcommand{\HBTrained}{67.6}
\newcommand{\HBGain}{\ensuremath{+12.8}}
\newcommand{\HBCI}{\ensuremath{[+9.5, +16.2]}}
\newcommand{\HBReleaseGain}{\ensuremath{+13.1}}
\newcommand{\HBReleaseCI}{\ensuremath{[+9.9, +16.6]}}
\newcommand{\ChartBase}{39.4}
\newcommand{\ChartTrained}{50.6}
\newcommand{\ChartGain}{\ensuremath{+11.2}}
\newcommand{\ChartCI}{\ensuremath{[+7.5, +15.0]}}
\newcommand{\CMOfficialBase}{76.90}
\newcommand{\CMOfficialTrained}{84.25}
\newcommand{\CMOfficialGain}{\ensuremath{+7.35}}
\newcommand{\CMOfficialCI}{\ensuremath{[+5.55, +9.15]}}
\newcommand{\CMLenientBase}{79.35}
\newcommand{\CMLenientTrained}{85.80}
\newcommand{\CMLenientGain}{\ensuremath{+6.45}}
\newcommand{\CMLenientCI}{\ensuremath{[+4.80, +8.10]}}
\newcommand{\GDPOfficialBase}{10.4}
\newcommand{\GDPOfficialTrained}{24.2}
\newcommand{\GDPOfficialGain}{\ensuremath{+13.8}}
\newcommand{\GDPOfficialCI}{\ensuremath{[+7.8, +20.2]}}
\newcommand{\GDPCorrectedBase}{12.4}
\newcommand{\GDPCorrectedTrained}{31.0}
\newcommand{\GDPCorrectedGain}{\ensuremath{+18.6}}
\newcommand{\GDPCorrectedCI}{\ensuremath{[+12.2, +25.0]}}
\newcommand{\SWEVerifiedBase}{73.3}
\newcommand{\SWEVerifiedTrained}{72.8}
\newcommand{\SWEVerifiedGain}{\ensuremath{-0.5}}
\newcommand{\SWEVerifiedCI}{\ensuremath{[-2.7, +1.8]}}
\newcommand{\SWEHeldoutBase}{35.4}
\newcommand{\SWEHeldoutTrained}{37.8}
\newcommand{\SWEHeldoutGain}{\ensuremath{+2.4}}
\newcommand{\SWEHeldoutCI}{\ensuremath{[-0.2, +5.1]}}
\newcommand{\TBenchRunTwoBase}{63.8}
\newcommand{\TBenchRunTwoTrained}{69.4}
\newcommand{\TBenchRunTwoGain}{\ensuremath{+5.6}}
\newcommand{\TBenchRunTwoCI}{\ensuremath{[+2.0, +9.4]}}
\newcommand{\SWEProRunTwoBase}{86.6}
\newcommand{\SWEProRunTwoTrained}{91.8}
\newcommand{\SWEProRunTwoGain}{\ensuremath{+5.2}}
\newcommand{\SWEProRunTwoCI}{\ensuremath{[+3.3, +7.1]}}
\newcommand{\TBenchTerminusBase}{65.8}
\newcommand{\TBenchTerminusTrained}{67.4}
\newcommand{\TBenchTerminusGain}{\ensuremath{+1.6}}
\newcommand{\TBenchTerminusCI}{\ensuremath{[-2.2, +5.4]}}
\newcommand{\DeepSWEFullMini}{10.7}
\newcommand{\DeepSWEVTwo}{27.5}
\newcommand{\CharXivBase}{84.05}
\newcommand{\CharXivTrained}{85.45}
\newcommand{\CharXivGain}{\ensuremath{+1.40}}
\newcommand{\CharXivCI}{\ensuremath{[-0.05, +2.85]}}
\newcommand{\CharXivNoToolsBase}{77.05}
\newcommand{\CharXivNoToolsTrained}{78.25}
\newcommand{\CharXivNoToolsGain}{\ensuremath{+1.20}}
\newcommand{\CharXivNoToolsCI}{\ensuremath{[-0.25, +2.70]}}
\newcommand{\ChartNoToolsBase}{15.9}
\newcommand{\ChartNoToolsTrained}{16.1}
\newcommand{\ChartNoToolsGain}{\ensuremath{+0.2}}
\newcommand{\ChartNoToolsCI}{\ensuremath{[-2.6, +2.8]}}
\newcommand{\ChartNoToolsNativeGain}{\ensuremath{+0.5}}
\newcommand{\ChartNoToolsNativeCI}{\ensuremath{[-1.8, +2.7]}}
\newcommand{\CMNoToolsLenientBase}{70.50}
\newcommand{\CMNoToolsLenientTrained}{72.25}
\newcommand{\CMNoToolsLenientGain}{\ensuremath{+1.75}}
\newcommand{\CMNoToolsLenientCI}{\ensuremath{[+0.30, +3.20]}}
\newcommand{\ChartQAProBase}{68.79}
\newcommand{\ChartQAProTrained}{68.89}
\newcommand{\ChartQAProGain}{\ensuremath{+0.10}}
\newcommand{\ChartQAProCI}{\ensuremath{[-0.69, +0.92]}}
\newcommand{\OCRReasoningBase}{66.18}
\newcommand{\OCRReasoningTrained}{65.90}
\newcommand{\OCRReasoningGain}{\ensuremath{-0.28}}
\newcommand{\OCRReasoningCI}{\ensuremath{[-1.68, +1.12]}}
\newcommand{\OCRBenchBase}{68.32}
\newcommand{\OCRBenchTrained}{68.57}
\newcommand{\OCRBenchGain}{\ensuremath{+0.25}}
\newcommand{\OCRBenchCI}{\ensuremath{[-0.29, +0.79]}}
\newcommand{\ChartRerunBase}{36.0}
\newcommand{\ChartRerunTrained}{47.3}
\newcommand{\ChartRerunGain}{\ensuremath{+11.3}}
\newcommand{\ChartRerunCI}{\ensuremath{[+5.0, +17.7]}}
\newcommand{\ChartRerunAttempts}{3}
\newcommand{\GDPNoToolsGain}{\ensuremath{+3.6}}
\newcommand{\GDPNoToolsCI}{\ensuremath{[+0.6, +6.8]}}
\newcommand{\GDPNoToolsFirstGain}{\ensuremath{+3.4}}
\newcommand{\GDPNoToolsFirstCI}{\ensuremath{[-0.2, +7.6]}}
\newcommand{\GDPTempSixBase}{10.2}
\newcommand{\GDPTempSixTrained}{19.6}
\newcommand{\GDPTempSixGain}{\ensuremath{+9.4}}
\newcommand{\GDPTempSixCI}{\ensuremath{[+4.0, +15.0]}}
\newcommand{\GDPNoToolsFirstBase}{7.0}
\newcommand{\GDPNoToolsFirstTrained}{10.4}
\newcommand{\GDPLowTemperature}{0.6}
\newcommand{\SWEHeldoutMediumBase}{31.1}
\newcommand{\SWEHeldoutMediumTrained}{39.7}
\newcommand{\SWEHeldoutMediumGain}{\ensuremath{+8.6}}
\newcommand{\SWEHeldoutMediumCI}{\ensuremath{[+6.4, +10.8]}}
\newcommand{\SWEProGoBase}{87.9}
\newcommand{\SWEProGoTrained}{93.1}
\newcommand{\SWEProGoGain}{\ensuremath{+5.1}}
\newcommand{\SWEProGoCI}{\ensuremath{[+2.4, +8.1]}}
\newcommand{\SWEProGoBWT}{32/10/211}
\newcommand{\SWEProPythonBase}{87.9}
\newcommand{\SWEProPythonTrained}{92.6}
\newcommand{\SWEProPythonGain}{\ensuremath{+4.7}}
\newcommand{\SWEProPythonCI}{\ensuremath{[+1.9, +7.7]}}
\newcommand{\SWEProPythonBWT}{24/8/203}
\newcommand{\SWEProTypeScriptBase}{83.2}
\newcommand{\SWEProTypeScriptTrained}{89.5}
\newcommand{\SWEProTypeScriptGain}{\ensuremath{+6.4}}
\newcommand{\SWEProTypeScriptCI}{\ensuremath{[+1.4, +11.8]}}
\newcommand{\SWEProTypeScriptBWT}{17/7/86}
\newcommand{\SWEProJavaScriptBase}{85.1}
\newcommand{\SWEProJavaScriptTrained}{90.5}
\newcommand{\SWEProJavaScriptGain}{\ensuremath{+5.4}}
\newcommand{\SWEProJavaScriptCI}{\ensuremath{[-5.4, +16.2]}}
\newcommand{\SWEProJavaScriptBWT}{7/3/27}
\newcommand{\CCImplicitBase}{13.4}
\newcommand{\CCImplicitTrained}{31.3}
\newcommand{\CCImplicitGain}{\ensuremath{+17.9}}
\newcommand{\CCImplicitCI}{\ensuremath{[+4.5, +33.0]}}
\newcommand{\CCImplicitBWT}{6/1/7}
\newcommand{\CCMultistepBase}{27.5}
\newcommand{\CCMultistepTrained}{42.5}
\newcommand{\CCMultistepGain}{\ensuremath{+15.0}}
\newcommand{\CCMultistepCI}{\ensuremath{[+2.5, +27.5]}}
\newcommand{\CCMultistepBWT}{10/3/7}
\newcommand{\CCNegativeBase}{31.3}
\newcommand{\CCNegativeTrained}{33.5}
\newcommand{\CCNegativeGain}{\ensuremath{+2.3}}
\newcommand{\CCNegativeCI}{\ensuremath{[-6.3, +11.4]}}
\newcommand{\CCNegativeBWT}{7/9/6}
\newcommand{\CCConditionalBase}{32.9}
\newcommand{\CCConditionalTrained}{33.6}
\newcommand{\CCConditionalGain}{\ensuremath{+0.7}}
\newcommand{\CCConditionalCI}{\ensuremath{[-11.2, +10.5]}}
\newcommand{\CCConditionalBWT}{7/2/10}
\newcommand{\IFBWordsBase}{57.0}
\newcommand{\IFBWordsTrained}{65.6}
\newcommand{\IFBWordsGain}{\ensuremath{+8.6}}
\newcommand{\IFBWordsCI}{\ensuremath{[+4.3, +12.9]}}
\newcommand{\IFBWordsBWT}{42/11/32}
\newcommand{\IFBFormatBase}{81.8}
\newcommand{\IFBFormatTrained}{86.1}
\newcommand{\IFBFormatGain}{\ensuremath{+4.2}}
\newcommand{\IFBFormatCI}{\ensuremath{[-0.2, +9.0]}}
\newcommand{\IFBFormatBWT}{23/14/59}
\newcommand{\IFBCountBase}{85.7}
\newcommand{\IFBCountTrained}{89.9}
\newcommand{\IFBCountGain}{\ensuremath{+4.2}}
\newcommand{\IFBCountCI}{\ensuremath{[-0.4, +8.8]}}
\newcommand{\IFBCountBWT}{21/12/29}
\newcommand{\IFBSentenceBase}{72.1}
\newcommand{\IFBSentenceTrained}{74.7}
\newcommand{\IFBSentenceGain}{\ensuremath{+2.6}}
\newcommand{\IFBSentenceCI}{\ensuremath{[-3.8, +9.3]}}
\newcommand{\IFBSentenceBWT}{5/4/19}
\newcommand{\IFBRatioBase}{67.4}
\newcommand{\IFBRatioTrained}{68.2}
\newcommand{\IFBRatioGain}{\ensuremath{+0.9}}
\newcommand{\IFBRatioCI}{\ensuremath{[-5.2, +6.4]}}
\newcommand{\IFBRatioBWT}{9/7/28}
\newcommand{\IFBRepeatBase}{41.7}
\newcommand{\IFBRepeatTrained}{41.7}
\newcommand{\IFBRepeatGain}{0.0}
\newcommand{\IFBRepeatCI}{\ensuremath{[-4.2, +4.2]}}
\newcommand{\IFBRepeatBWT}{1/1/7}
\newcommand{\IFBCustomBase}{78.4}
\newcommand{\IFBCustomTrained}{77.3}
\newcommand{\IFBCustomGain}{\ensuremath{-1.1}}
\newcommand{\IFBCustomCI}{\ensuremath{[-5.7, +3.4]}}
\newcommand{\IFBCustomBWT}{2/3/6}
\newcommand{\MCInferenceBase}{60.7}
\newcommand{\MCInferenceTrained}{70.0}
\newcommand{\MCInferenceGain}{\ensuremath{+9.3}}
\newcommand{\MCInferenceCI}{\ensuremath{[+4.9, +13.9]}}
\newcommand{\MCInferenceBWT}{46/16/47}
\newcommand{\MCRetentionBase}{74.8}
\newcommand{\MCRetentionTrained}{81.2}
\newcommand{\MCRetentionGain}{\ensuremath{+6.3}}
\newcommand{\MCRetentionCI}{\ensuremath{[+2.9, +10.3]}}
\newcommand{\MCRetentionBWT}{23/5/41}
\newcommand{\MCVersionEditingBase}{45.0}
\newcommand{\MCVersionEditingTrained}{59.4}
\newcommand{\MCVersionEditingGain}{\ensuremath{+14.4}}
\newcommand{\MCVersionEditingCI}{\ensuremath{[+6.3, +22.8]}}
\newcommand{\MCVersionEditingBWT}{19/6/15}
\newcommand{\MCSelfCoherenceBase}{52.6}
\newcommand{\MCSelfCoherenceTrained}{55.7}
\newcommand{\MCSelfCoherenceGain}{\ensuremath{+3.1}}
\newcommand{\MCSelfCoherenceCI}{\ensuremath{[-2.3, +8.3]}}
\newcommand{\MCSelfCoherenceBWT}{18/10/20}
\newcommand{\AIFCarriedBase}{53.1}
\newcommand{\AIFCarriedTrained}{60.2}
\newcommand{\AIFCarriedGain}{\ensuremath{+7.1}}
\newcommand{\AIFCarriedCI}{\ensuremath{[+4.8, +9.6]}}
\newcommand{\AIFCarriedBWT}{168/80/488}
\newcommand{\AIFComplexBase}{59.2}
\newcommand{\AIFComplexTrained}{66.8}
\newcommand{\AIFComplexGain}{\ensuremath{+7.6}}
\newcommand{\AIFComplexCI}{\ensuremath{[+3.6, +11.7]}}
\newcommand{\AIFComplexBWT}{107/61/234}
\newcommand{\AIFSteerBase}{39.5}
\newcommand{\AIFSteerTrained}{48.0}
\newcommand{\AIFSteerGain}{\ensuremath{+8.5}}
\newcommand{\AIFSteerCI}{\ensuremath{[+5.1, +11.9]}}
\newcommand{\AIFSteerBWT}{140/65/300}
\newcommand{\ABHRBase}{10.0}
\newcommand{\ABHRTrained}{33.0}
\newcommand{\ABHRGain}{\ensuremath{+23.0}}
\newcommand{\ABHRCI}{\ensuremath{[+15.0, +32.0]}}
\newcommand{\ABHRBWT}{24/1/75}
\newcommand{\ABSalesBase}{7.0}
\newcommand{\ABSalesTrained}{22.0}
\newcommand{\ABSalesGain}{\ensuremath{+15.0}}
\newcommand{\ABSalesCI}{\ensuremath{[+7.0, +23.0]}}
\newcommand{\ABSalesBWT}{17/2/81}
\newcommand{\ABOperationsBase}{24.0}
\newcommand{\ABOperationsTrained}{37.0}
\newcommand{\ABOperationsGain}{\ensuremath{+13.0}}
\newcommand{\ABOperationsCI}{\ensuremath{[+4.0, +22.0]}}
\newcommand{\ABOperationsBWT}{18/5/77}
\newcommand{\ABFinanceBase}{17.0}
\newcommand{\ABFinanceTrained}{27.0}
\newcommand{\ABFinanceGain}{\ensuremath{+10.0}}
\newcommand{\ABFinanceCI}{\ensuremath{[+1.0, +19.0]}}
\newcommand{\ABFinanceBWT}{16/6/78}
\newcommand{\ABMarketingBase}{11.0}
\newcommand{\ABMarketingTrained}{21.0}
\newcommand{\ABMarketingGain}{\ensuremath{+10.0}}
\newcommand{\ABMarketingCI}{\ensuremath{[+2.0, +18.0]}}
\newcommand{\ABMarketingBWT}{14/4/82}
\newcommand{\ABSupportBase}{20.0}
\newcommand{\ABSupportTrained}{25.0}
\newcommand{\ABSupportGain}{\ensuremath{+5.0}}
\newcommand{\ABSupportCI}{\ensuremath{[-2.0, +13.0]}}
\newcommand{\ABSupportBWT}{10/5/85}
\newcommand{\HBHRBase}{49.0}
\newcommand{\HBHRTrained}{69.0}
\newcommand{\HBHRGain}{\ensuremath{+20.0}}
\newcommand{\HBHRCI}{\ensuremath{[+14.3, +26.0]}}
\newcommand{\HBHRBWT}{13/0/0}
\newcommand{\HBLogisticsBase}{43.8}
\newcommand{\HBLogisticsTrained}{59.3}
\newcommand{\HBLogisticsGain}{\ensuremath{+15.5}}
\newcommand{\HBLogisticsCI}{\ensuremath{[+8.1, +22.0]}}
\newcommand{\HBLogisticsBWT}{11/1/0}
\newcommand{\HBFinanceBase}{63.4}
\newcommand{\HBFinanceTrained}{74.2}
\newcommand{\HBFinanceGain}{\ensuremath{+10.8}}
\newcommand{\HBFinanceCI}{\ensuremath{[+4.2, +17.4]}}
\newcommand{\HBFinanceBWT}{8/3/1}
\newcommand{\HBInsuranceBase}{57.7}
\newcommand{\HBInsuranceTrained}{68.2}
\newcommand{\HBInsuranceGain}{\ensuremath{+10.5}}
\newcommand{\HBInsuranceCI}{\ensuremath{[+2.3, +20.6]}}
\newcommand{\HBInsuranceBWT}{9/2/2}
\newcommand{\HBMedicalBase}{59.2}
\newcommand{\HBMedicalTrained}{67.1}
\newcommand{\HBMedicalGain}{\ensuremath{+7.9}}
\newcommand{\HBMedicalCI}{\ensuremath{[+1.9, +14.6]}}
\newcommand{\HBMedicalBWT}{10/5/0}
\newcommand{\ChartSTEMBase}{39.5}
\newcommand{\ChartSTEMTrained}{47.5}
\newcommand{\ChartSTEMGain}{\ensuremath{+8.0}}
\newcommand{\ChartSTEMCI}{\ensuremath{[+3.9, +12.5]}}
\newcommand{\ChartSTEMBWT}{29/8/24}
\newcommand{\ChartManufacturingBase}{28.9}
\newcommand{\ChartManufacturingTrained}{51.6}
\newcommand{\ChartManufacturingGain}{\ensuremath{+22.6}}
\newcommand{\ChartManufacturingCI}{\ensuremath{[+13.7, +32.1]}}
\newcommand{\ChartManufacturingBWT}{14/1/4}
\newcommand{\ChartHealthcareBase}{42.7}
\newcommand{\ChartHealthcareTrained}{49.1}
\newcommand{\ChartHealthcareGain}{\ensuremath{+6.4}}
\newcommand{\ChartHealthcareCI}{\ensuremath{[-5.5, +21.8]}}
\newcommand{\ChartHealthcareBWT}{4/2/5}
\newcommand{\ChartFinanceBase}{56.7}
\newcommand{\ChartFinanceTrained}{71.1}
\newcommand{\ChartFinanceGain}{\ensuremath{+14.4}}
\newcommand{\ChartFinanceCI}{\ensuremath{[+4.4, +25.6]}}
\newcommand{\ChartFinanceBWT}{5/0/4}
\newcommand{\IFBLineIndentBase}{56.9}
\newcommand{\IFBLineIndentTrained}{33.3}
\newcommand{\IFBLineIndentGain}{\ensuremath{-23.6}}
\newcommand{\IFBLineIndentCI}{\ensuremath{[-36.1, -9.7]}}
\newcommand{\XferPassBase}{13.5}
\newcommand{\XferPassTrained}{21.5}
\newcommand{\XferPassGain}{\ensuremath{+8.0}}
\newcommand{\XferPassCI}{\ensuremath{[+5.2, +11.0]}}
\newcommand{\XferOutcomeGain}{\ensuremath{+7.8}}
\newcommand{\XferOutcomeCI}{\ensuremath{[+4.7, +10.9]}}
\newcommand{\XferChecksGain}{\ensuremath{+15.4}}
\newcommand{\XferChecksCI}{\ensuremath{[+12.5, +18.2]}}
\newcommand{\XferRuleBreakBase}{34.7}
\newcommand{\XferRuleBreakTrained}{40.0}
\newcommand{\XferRuleBreakGain}{\ensuremath{+5.3}}
\newcommand{\XferRuleBreakCI}{\ensuremath{[+1.7, +9.0]}}
\newcommand{\XferOnlyTrained}{69}
\newcommand{\DocsABNeverInboxGain}{\ensuremath{-14.7}}
\newcommand{\DocsABNeverInboxCI}{\ensuremath{[-18.2, -11.2]}}
\newcommand{\DocsABNeverInboxBase}{71.5}
\newcommand{\DocsABNeverInboxTrained}{56.8}
\newcommand{\BehNeverInboxBase}{74.3}
\newcommand{\BehNeverInboxTrained}{41.2}
\newcommand{\BehNeverInboxGain}{\ensuremath{-33.2}}
\newcommand{\BehNeverInboxCI}{\ensuremath{[-37.0, -29.5]}}
\newcommand{\BehMsgSeedBase}{26.6}
\newcommand{\BehMsgSeedTrained}{67.9}
\newcommand{\BehMsgSeedGain}{\ensuremath{+41.2}}
\newcommand{\BehMsgSeedCI}{\ensuremath{[+34.1, +48.4]}}
\newcommand{\BehTestsAfterEditBase}{83.9}
\newcommand{\BehTestsAfterEditTrained}{91.9}
\newcommand{\BehTestsAfterEditGain}{\ensuremath{+8.0}}
\newcommand{\BehTestsAfterEditCI}{\ensuremath{[+5.7, +10.2]}}
\newcommand{\BehNoMeasurementBase}{38.7}
\newcommand{\BehNoMeasurementTrained}{12.0}
\newcommand{\BehNoMeasurementGain}{\ensuremath{-26.7}}
\newcommand{\BehNoMeasurementCI}{\ensuremath{[-34.0, -19.7]}}
\newcommand{\TokSWEProRatio}{2.07}
\newcommand{\TokCMRatio}{1.23}
\newcommand{\GDPIntentGain}{\ensuremath{+18.9}}
\newcommand{\GDPIntentCI}{\ensuremath{[+14.4, +23.7]}}
\newcommand{\GDPNoToolsCheckBase}{10.4}
\newcommand{\GDPNoToolsCheckTrained}{14.2}
\newcommand{\GDPNoToolsPromptGain}{\ensuremath{+5.0}}
\newcommand{\GDPNoToolsPromptCI}{\ensuremath{[+1.4, +9.0]}}
\newcommand{\GDPNoToolsCheckGain}{\ensuremath{+3.8}}
\newcommand{\GDPNoToolsCheckCI}{\ensuremath{[-0.4, +7.8]}}
\newcommand{\CCDevBase}{40.2}
\newcommand{\CCDevTrained}{51.8}
\newcommand{\RecipeMaxTokensEarlyConstraints}{32,768}
\newcommand{\SxsSampleSize}{30}
\newcommand{\BehNeverSheetsBase}{24.4}
\newcommand{\BehNeverSheetsTrained}{7.2}
\newcommand{\BehNeverSheetsGain}{\ensuremath{-17.2}}
\newcommand{\BehNeverSheetsCI}{\ensuremath{[-20.6, -14.0]}}
\newcommand{\SxsToolUsePairsRead}{30}
\newcommand{\SxsToolUseModeNotFound}{12}
\newcommand{\SxsToolUseModeMissedRule}{10}
\newcommand{\SxsToolUseModeStep}{7}
\newcommand{\SxsToolUseModeUnclassified}{1}
\newcommand{\SxsBlindToolUseSame}{29}
\newcommand{\PairSWERead}{30}
\newcommand{\PairSWEModeChecks}{8}
\newcommand{\PairSWEModeLostReq}{13}
\newcommand{\PairSWEModeBroke}{5}
\newcommand{\PairSWEModeUnclear}{4}
\newcommand{\PairSWEProOwnFileBase}{14.3}
\newcommand{\PairSWEProOwnFileTrained}{32.6}
\newcommand{\PairSWEProOwnFileGain}{\ensuremath{+18.3}}
\newcommand{\PairSWEProOwnFileCI}{\ensuremath{[+15.7, +20.9]}}
\newcommand{\PairDeepSWEOwnFileBase}{82.6}
\newcommand{\PairDeepSWEOwnFileTrained}{85.9}
\newcommand{\SxsBlindSWESame}{23}
\newcommand{\PairIFBNewlySolved}{3}
\newcommand{\PairMCNewlySolved}{8}
\newcommand{\PairPlanBaseNoBP}{4}
\newcommand{\PairPlanTrainedPasses}{5}
\newcommand{\PairIFBBoldBase}{17.1}
\newcommand{\PairIFBBoldTrained}{5.8}
\newcommand{\PairIFBBoldGain}{\ensuremath{-11.3}}
\newcommand{\PairIFBBoldCI}{\ensuremath{[-14.3, -8.5]}}
\newcommand{\PairIndentAttempts}{72}
\newcommand{\PairIndentFenceTrained}{48}
\newcommand{\PairIndentFenceBase}{30}
\newcommand{\SxsConstraintsPairsRead}{11}
\newcommand{\PairMCNoticedBase}{18}
\newcommand{\PairMCNoticedAttempts}{24}
\newcommand{\BehCMNoMeasurementVisualBase}{67.9}
\newcommand{\BehCMNoMeasurementVisualTrained}{44.4}
\newcommand{\BehCMNoMeasurementVisualGain}{\ensuremath{-23.5}}
\newcommand{\BehCMNoMeasurementVisualCI}{\ensuremath{[-25.9, -21.2]}}
\newcommand{\SxsCMNewlySolved}{24}
\newcommand{\SxsCMPoolAttempts}{48}
\newcommand{\SxsCMPoolBaseMeasured}{13}
\newcommand{\SxsCMPoolTrainedMeasured}{37}
\newcommand{\SxsCMModeEye}{10}
\newcommand{\SxsCMModeEncoding}{7}
\newcommand{\SxsCMModePartial}{6}
\newcommand{\SxsCMModeOther}{1}
\newcommand{\ExDocsReadPairs}{30}
\newcommand{\ExDocsModeSourceCount}{11}
\newcommand{\ExDocsModeDetailCount}{11}
\newcommand{\ExDocsModeMessagesCount}{6}
\newcommand{\ExDocsModeStopCount}{2}
\newcommand{\DocsABNeverSheetBase}{27.5}
\newcommand{\DocsABNeverSheetTrained}{11.7}
\newcommand{\DocsABNeverSheetGain}{\ensuremath{-15.7}}
\newcommand{\DocsABNeverSheetCI}{\ensuremath{[-19.1, -12.3]}}

\title{What RL on Expert-Built Tasks Teaches\\Nemotron~3.5~Super}

\author{Sushant Mehta \qquad Edwin Chen\\
  Surge AI\\
  \texttt{research@surgehq.ai}
}

\begin{document}

\maketitle

\begin{abstract}
We post-train NVIDIA's Nemotron~3.5 Super (\ModelTotalParams{}B total, \ModelActiveParams{}B active parameters) with reinforcement learning (RL) on five expert-built task datasets, one at a time: real-world software engineering, complex constraints, long-horizon tool use, professional chart understanding and enterprise document reasoning. Each run trains a fresh rank-\RecipeLoRARankSWE{} LoRA adapter with CISPO on Tinker, with no supervised stage, and each adapter is compared with the base model on the same tasks, harness and settings. On twelve headline benchmarks, eight built by other groups and four by Surge AI, the gains range from \IFBGain{} points on IFBench to \GDPOfficialGain{} on GDP.pdf, which rises from \GDPOfficialBase\% to \GDPOfficialTrained\%; SWE-bench Pro V2 rises from \SWEProBase\% to \SWEProTrained\%, Terminal-Bench~2.1 from \TBenchBase\% to \TBenchTrained\%, MultiChallenge from \MCBase\% to \MCTrained\%, AutomationBench from \ABBase\% to \ABTrained\% and Chartography from \ChartBase\% to \ChartTrained\%. The document adapter, trained without API tools, also gains \XferPassGain{} points on AutomationBench. In attempts read side by side with the base model's on tasks each adapter newly solves, the base model's failures recur in a few modes, and each adapter does more often what the base model missed: the tool-use adapter finds the data and the rules before acting, the coding adapter writes its own test or scratch files on SWE-bench Pro~V2, the constraints adapter carries earlier turns into its answers, the chart adapter measures the marks in code, and the document adapter reads AutomationBench's spreadsheets.
\end{abstract}
\section{Introduction}
\label{sec:intro}

Reinforcement learning (RL) on tasks with verifiable or rubric-graded rewards is now a standard stage of post-training~\citep{deepseekr1,tulu3}. Most published recipes train on a mixture of datasets, so they show what the mixture adds rather than what one dataset adds to a strong open model: on which benchmarks it helps, by how much, and through which changes in~behavior.

We answer these questions for NVIDIA's Nemotron~3.5 Super~\citep{nemotron35,nemotron35card}, a hybrid Mamba-Transformer mixture-of-experts model with \ModelTotalParams{}B total and \ModelActiveParams{}B active parameters. We train it with RL on five of Surge AI's expert-built task datasets, one at a time: Real-World SWE, Complex Constraints, Long-Horizon Tool Use, Professional Chart Understanding and Enterprise Document Reasoning (\S\ref{sec:training}). Each run trains a fresh rank-\RecipeLoRARankSWE{} LoRA adapter~\citep{lora} with CISPO~\citep{minimaxm1} on Tinker~\citep{tinker}, with no supervised stage and no KL penalty. We call the five adapters RL-SWE, RL-Constraints, RL-ToolUse, RL-Charts and RL-Docs. Each is compared with the base model on the same tasks, in the same harness and with the same settings, and every gain carries a paired 95\% task-bootstrap interval (\S\ref{sec:eval}).

\begin{figure}[t]
  \centering
  \includegraphics[width=\textwidth]{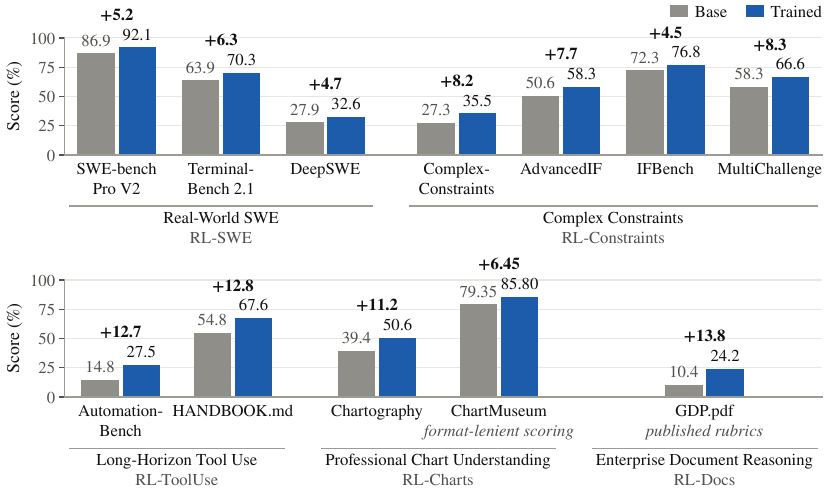}
  \caption{Base and trained scores on the twelve headline benchmarks, grouped by the adapter's training dataset. Bold numbers are gains in percentage points. Metrics as in Table~\ref{tab:headline}.}
  \label{fig:headline}
\end{figure}

Over the twelve headline benchmarks (Figure~\ref{fig:headline}, \S\ref{sec:results}), eight built by other groups and four by Surge AI, the gains range from \IFBGain{} points on IFBench to \GDPOfficialGain{} on GDP.pdf, where RL-Docs lifts the score from \GDPOfficialBase\% to \GDPOfficialTrained\%. RL-SWE lifts SWE-bench Pro V2 from \SWEProBase\% to \SWEProTrained\%. RL-ToolUse raises AutomationBench's pass rate from \ABBase\% to \ABTrained\%. RL-Constraints, trained only on single-turn requests, gains \MCGain{} points on MultiChallenge, a multi-turn benchmark. RL-Docs, trained on questions about PDF documents with no API tools, also gains \XferPassGain{} points on AutomationBench.

To see what changed, we read base and trained attempts side by side on public-benchmark tasks each adapter newly solves, grouped base's failures into modes and measured matching behaviors over all attempts (\S\ref{sec:behavior}). On AutomationBench, base often acted without finding the task's data or the rule governing it; RL-ToolUse looks first: the share of tasks where it never opens the inbox or Slack falls from \BehNeverInboxBase\% to \BehNeverInboxTrained\%. On the coding benchmarks, base often dropped a requirement or ran checks that missed its error; on SWE-bench Pro~V2, RL-SWE writes its own test or scratch files more often. On MultiChallenge, base answered the latest turn but dropped what an earlier turn set, which RL-Constraints carries into its answers; on ChartMuseum, base judged close calls by eye; RL-Charts measures the marks in code. RL-Docs reads AutomationBench's spreadsheets more~often.

\paragraph{Contributions.}
\begin{enumerate}[leftmargin=*,itemsep=1pt,topsep=2pt]
  \item Five RL runs on one base model, one per expert-built dataset, with a shared recipe: CISPO on rank-\RecipeLoRARankSWE{} LoRA adapters, with no supervised stage and no KL penalty (\S\ref{sec:training}).
  \item A paired evaluation of each adapter against base on twelve headline benchmarks, on the same tasks and harness and with intervals, plus every other evaluation of the same checkpoints (\S\S\ref{sec:eval}--\ref{sec:results}).
  \item A side-by-side analysis that names the leading base failure modes, measures matching behaviors over all attempts, and links each mode to the training reward (\S\ref{sec:behavior}).
\end{enumerate}
\section{Related Work}
\label{sec:related}

\paragraph{RL post-training for agents and instruction following.} Rule-based rewards drove large-scale reasoning RL in DeepSeek-R1~\citep{deepseekr1}, and T\"ulu~3~\citep{tulu3} made RL with verifiable rewards a stage of an open post-training recipe. GRPO~\citep{deepseekmath} introduced group-relative advantages; Dr.~GRPO~\citep{drgrpo} removed the per-group standard-deviation scaling, as our runs do; CISPO~\citep{minimaxm1} clips the importance weight instead of the update, and ScaleRL~\citep{scalerl} builds a large-scale recipe on it. For software-engineering agents, SWE-RL~\citep{swerl} learns from software-evolution data with a patch-similarity reward, and DeepSWE-Preview~\citep{deepsweagent}, SkyRL~\citep{skyrl} and \citet{golubev2025} train with RL in executable repository environments, which SWE-Gym~\citep{swegym} and SWE-smith~\citep{swesmith} build at scale. \citet{lorawithoutregret} find that LoRA applied to every weight matrix, the MLP layers included, matches full fine-tuning for RL at small ranks. For the same base model, NVIDIA describes its own post-training~\citep{nemotron35}.

\paragraph{Rubric and verifiable rewards.} RL rewards have moved from exact answers to instruction following and open-ended tasks through verifiable constraints~\citep{ifbench}, checklists~\citep{checklists} and instance-specific rubrics~\citep{rubricsasrewards,advancedif,complexconstraints}. Four of our five rewards are graded by a model judge against a rubric or an expert answer, and one by hidden tests. The rubrics state individually gradable requirements, and two of the rewards add a bonus for meeting all of them (\S\ref{sec:data}).

\paragraph{Transfer from post-training.} Studies of RL across reasoning domains find that what transfers between domains depends on the domains~\citep{guru,onedomain}. Prior work from our group reports transfer across benchmarks and harnesses from RL on expert-built agentic coding tasks~\citep{crossbenchcoding}, gains in software engineering from post-training on long-horizon office work~\citep{crossbenchlonghorizon,officetransfer}, and multi-turn gains from single-turn constraint training~\citep{complexconstraints}. Here each run starts from Nemotron~3.5 Super and the dataset varies, so each adapter's gains and behaviors can be read against its own training data.
\section{Training}
\label{sec:training}

\subsection{Five datasets}
\label{sec:data}

Each run trains on one of Surge AI's expert-built task datasets (Table~\ref{tab:data}). No run uses reference trajectories, supervised fine-tuning or distillation.

\begin{table}[t]
  \caption{The five training datasets. $c$ is the share of a rollout's rubric criteria met and $a$ is 1 when all are met. Every judge reads only the final visible answer or the final state, never the reasoning.}
  \label{tab:data}
  \centering
  \footnotesize
  \setlength{\tabcolsep}{3pt}
  \begin{tabularx}{\linewidth}{@{}>{\raggedright\arraybackslash}p{2.75cm} >{\raggedright\arraybackslash}p{1.45cm} Y Y Y@{}}
    \toprule
    Dataset (adapter) & Domain & Task format & Reward and grader & Training harness \\
    \midrule
    Real-World SWE (RL-SWE) & Software engineering & A repository at a fixed commit and an expert-written change request & Pass only if every hidden test passes & OpenCode and mini-swe-agent ports, network blocked \\
    \addlinespace[2pt]
    Complex Constraints (RL-Constraints) & Instruction following & A single-turn request, often with source material, and a binary rubric & $(c + w\,a)/(1 + w)$, $w=\RecipeRewardAllPassWeightConstraints$; GPT-6.1 Sol judge & One response, no tools \\
    \addlinespace[2pt]
    Long-Horizon Tool Use (RL-ToolUse) & Agentic tool use & A request inside a simulated company, with a rubric of requirements and prohibitions & Share of requirements met, cut for broken prohibitions; GPT-6.1 Sol judge & The company's services as MCP tools, network blocked \\
    \addlinespace[2pt]
    Professional Chart Understanding (RL-Charts) & Chart reading & A professional chart, a value-reading question and an expert answer & Pass only if judged correct; GPT-6.1 Sol judge & Python kernel and crop tool \\
    \addlinespace[2pt]
    Enterprise Document Reasoning (RL-Docs) & Document reasoning & A PDF, a question and a rubric whose criteria carry expected values & $(c + w\,a)/(1 + w)$, $w=\RecipeRewardDocs$; GPT-6.1 Sol judge & Page-view, zoom, search and Python tools \\
    \bottomrule
  \end{tabularx}
\end{table}

\paragraph{Real-World SWE.} Each task pairs a real open-source repository at a fixed commit with an expert-written request for a feature, a bug fix or a behavior change, graded by hidden fail-to-pass and pass-to-pass tests in the format of SWE-bench~\citep{swebench}. A rollout earns the reward only if every hidden test passes in a fresh grading sandbox. The rollout groups ran in two harnesses, a port of OpenCode~\citep{opencode} and a port of mini-swe-agent~\citep{minisweagent,sweagent}, and every task in a repository of SWE-bench Verified, SWE-bench Pro or DeepSWE was removed before training.

\paragraph{Complex Constraints.} Each task is one professional or personal request with an expert-written rubric of binary criteria, in the format of the ComplexConstraints benchmark~\citep{complexconstraints}. GPT-6.1 Sol grades all criteria of the visible answer in one call, and the reward adds a bonus for meeting every one, so most of its spread within a group comes from that bonus. The judge's instructions say that a response's own claims about meeting a criterion are not evidence.

\paragraph{Long-Horizon Tool Use.} Each task places the agent inside a simulated company whose services, among them mail, chat, calendar, tickets and files, are exposed as Model Context Protocol (MCP) tools. A rubric lists what the request requires and what must not change. A GPT-6.1 Sol judge grades the services' final state and the agent's reply; the reward is the share of requirements met, and a change to something that must not change, or an unrequested side effect, drops it to zero or~below.

\paragraph{Professional Chart Understanding.} Each task pairs one real-world professional chart with a question that requires reading values off it and an expert answer, often an accepted range. The agent can run code in a persistent Python kernel and crop and enlarge regions of the chart. A rollout earns the reward only if a GPT-6.1 Sol judge, using the Chartography benchmark's own prompt, marks its final answer correct; hedged answers, such as two alternative values, score zero. Charts matching a Chartography chart were removed before training.

\paragraph{Enterprise Document Reasoning.} Each task is one PDF, a question and an expert-written rubric whose criteria carry their own expected values. The agent sees a first view of the document and can call page-view, zoom, search and Python tools. GPT-6.1 Sol grades one criterion per call with the GDP.pdf benchmark's own prompt, and training tasks whose document matched a GDP.pdf document were removed before training.

\subsection{Recipe}
\label{sec:recipe}

All five runs share one recipe; Appendix~\ref{app:training} lists the settings that differ. Each run trains a fresh rank-\RecipeLoRARankSWE{} LoRA adapter on the attention and MLP weights, the experts included, and the unembedding; the vision encoder is frozen in the chart and document runs. The loss is Tinker's CISPO~\citep{minimaxm1,tinkerdocs}: a token-level policy gradient whose importance weight, the ratio of trainer to sampler probabilities, is clipped to $[\RecipeClipLow, \RecipeClipHigh]$ and not differentiated. A rollout's advantage is its reward minus the mean reward of its group, with no standard-deviation scaling~\citep{drgrpo}, and groups whose rollouts all earn the same reward are dropped. There is no supervised stage, no KL penalty and no entropy bonus.

The optimizer is AdamW~\citep{adamw} with a peak learning rate of \mbox{\RecipeLRConstraints{}} (\mbox{\RecipeLRSWE{}} for RL-SWE) after a short warm-up. Rollouts sample at temperature \RecipeTrainTemperature{} and top-$p$ \RecipeTrainTopP{}, so that sampler probabilities match the policy the loss assumes. Sampling runs asynchronously: a group may start on sampler weights a few versions older than the trainer's, up to the run's staleness cap. Three runs add small shaping terms against overlong responses, cut-off turns or over-persistence.

Tinker served the base model from two deployments: RL-SWE and RL-ToolUse trained on an early-access deployment and the other three adapters on the deployment listed in Tinker's documentation, and each adapter is compared with base on its own deployment.
\section{Evaluation}
\label{sec:eval}

\begin{table}[t]
  \caption{The twelve headline benchmarks. Tasks are those scored for both models, times attempts per task per model. \textsuperscript{T}~marks the harness the adapter trained in. AdvancedIF's prompts and rubrics were written by Surge experts for Meta.}
  \label{tab:benchmarks}
  \centering
  \footnotesize
  \setlength{\tabcolsep}{3pt}
  \begin{tabularx}{\linewidth}{@{}l l l Y Y Y@{}}
    \toprule
    Benchmark & Adapter & Tasks & Metric & Grader & Harness \\
    \midrule
    \multicolumn{6}{@{}l}{\textit{Built by other groups}} \\
    SWE-bench Pro V2 & RL-SWE & \SizeSWEPro{} $\times$ \SWEProAttempts & Resolve rate & Hidden tests & OpenCode\textsuperscript{T}, no network \\
    Terminal-Bench 2.1 & RL-SWE & \SizeTBench{} $\times$ \TBenchAttempts & Pass rate & Hidden tests & OpenCode\textsuperscript{T}, open internet \\
    DeepSWE & RL-SWE & \SizeDeepSWE{} $\times$ \DeepSWEAttempts & Pass rate of the committed changes & Hidden tests & OpenCode\textsuperscript{T}, context compaction \\
    AdvancedIF & RL-Constraints & \SizeAIF{} $\times$ \AIFAttempts & Every rubric item met, mean of three subsets & Claude Opus 5.5 & One response\textsuperscript{T} per turn \\
    IFBench & RL-Constraints & \SizeIFB{} $\times$ \IFBAttempts & Share of constraints met, strict & Programmatic checkers & One response\textsuperscript{T} \\
    MultiChallenge & RL-Constraints & \SizeMC{} $\times$ \MCAttempts & Mean of four axis pass rates & Claude Opus 5.5 & Conversation replay\textsuperscript{T} \\
    AutomationBench & RL-ToolUse & \SizeAB{} $\times$ \ABAttempts & Pass rate, every assertion met & Programmatic assertions & Its own agent loop and API tools \\
    ChartMuseum & RL-Charts & \SizeCM{} $\times$ \CMAttempts & Accuracy, format-lenient & GPT-6.1 Sol & Python and crop tools\textsuperscript{T} \\
    \midrule
    \multicolumn{6}{@{}l}{\textit{Built by Surge AI}} \\
    ComplexConstraints & RL-Constraints & \SizeCC{} $\times$ \CCAttempts & Every criterion met & Claude Opus 5.5, per criterion & One response\textsuperscript{T} \\
    HANDBOOK.md v1.1 & RL-ToolUse & \SizeHB{} $\times$ \HBAttempts & Mean share of criteria met & GPT-6.1 Sol agent judge & OpenHands~\citep{openhands}, MCP tools \\
    Chartography & RL-Charts & \SizeChart{} $\times$ \ChartAttempts & Accuracy & GPT-6.1 Sol, official scorer & Python and crop tools\textsuperscript{T} \\
    GDP.pdf & RL-Docs & \SizeGDP{} $\times$ \GDPAttempts & Every criterion met & GPT-6.1 Sol, per criterion & Document tools\textsuperscript{T} \\
    \bottomrule
  \end{tabularx}
\end{table}

\paragraph{Benchmarks.} Table~\ref{tab:benchmarks} lists the twelve headline benchmarks~\citep{swebenchprov2,terminalbench,terminalbench21,deepswe,advancedif,ifbench,multichallenge,automationbench,chartmuseum,complexconstraints,handbook,chartography,gdppdf}. Each adapter is evaluated on the benchmarks of its own domain. Eight come from other groups, and ComplexConstraints, HANDBOOK.md, Chartography and GDP.pdf were built by Surge AI. Hidden tests or programmatic checkers grade the three coding benchmarks, IFBench and AutomationBench, and LLM judges grade the other seven. Section~\ref{sec:other} adds every other evaluation of the same checkpoints.

\paragraph{Paired protocol.} Base and each adapter run the same tasks in the same harness, with the same prompts, caps and sampling settings; only the adapter differs, and base runs on the adapter's own deployment. Every headline evaluation samples at temperature \EvalTemperature{} and the model's full reasoning effort, with top-$p$ \EvalTopP{}, or \ABTopP{} on AutomationBench and HANDBOOK.md as in training. Appendix~\ref{app:eval} gives the caps and the tasks scored.

\paragraph{Intervals.} Every gain is trained minus base in percentage points. Its 95\% interval is a paired task bootstrap~\citep{efron1979,miller2024}: per task we average each model's attempts and take the difference, then resample tasks with replacement \BootResamples{} times. Gains are computed from unrounded rates. Appendix~\ref{app:eval} gives the details.
\section{Results}
\label{sec:results}

\subsection{Headline benchmarks}
\label{sec:headline}

\begin{table}[t]
  \centering
  \begin{threeparttable}
  \caption{Headline results. Base and trained scores are in percent; the gain is trained minus base in percentage points, with its 95\% paired task-bootstrap interval. Metrics as in Table~\ref{tab:benchmarks}.}
  \label{tab:headline}
  \footnotesize
  \setlength{\tabcolsep}{5pt}
  \begin{tabular}{@{}l l r r r@{}}
    \toprule
    Benchmark & Adapter & Base & Trained & Gain [95\% CI] \\
    \midrule
    SWE-bench Pro V2 & RL-SWE & \SWEProBase & \SWEProTrained & \SWEProGain{}~\SWEProCI \\
    Terminal-Bench 2.1 & RL-SWE & \TBenchBase & \TBenchTrained & \TBenchGain{}~\TBenchCI \\
    DeepSWE & RL-SWE & \DeepSWEBase & \DeepSWETrained & \DeepSWEGain{}~\DeepSWECI \\
    \addlinespace[3pt]
    ComplexConstraints & RL-Constraints & \CCBase & \CCTrained & \CCGain{}~\CCCI \\
    AdvancedIF & RL-Constraints & \AIFBase & \AIFTrained & \AIFGain{}~\AIFCI \\
    IFBench & RL-Constraints & \IFBBase & \IFBTrained & \IFBGain{}~\IFBCI \\
    MultiChallenge & RL-Constraints & \MCBase & \MCTrained & \MCGain{}~\MCCI \\
    \addlinespace[3pt]
    AutomationBench & RL-ToolUse & \ABBase & \ABTrained & \ABGain{}~\ABCI \\
    HANDBOOK.md & RL-ToolUse & \HBBase & \HBTrained & \HBGain{}~\HBCI \\
    \addlinespace[3pt]
    Chartography & RL-Charts & \ChartBase & \ChartTrained & \ChartGain{}~\ChartCI \\
    ChartMuseum\tnote{a} & RL-Charts & \CMLenientBase & \CMLenientTrained & \CMLenientGain{}~\CMLenientCI \\
    \addlinespace[3pt]
    GDP.pdf, published rubrics & RL-Docs & \GDPOfficialBase & \GDPOfficialTrained & \GDPOfficialGain{}~\GDPOfficialCI \\
    \quad corrected rubrics\tnote{b} & RL-Docs & \GDPCorrectedBase & \GDPCorrectedTrained & \GDPCorrectedGain{}~\GDPCorrectedCI \\
    \bottomrule
  \end{tabular}
  \begin{tablenotes}[flushleft]\footnotesize
    \item[a] Format-lenient scoring, which reads the answer anywhere in the response. The official scorer, which reads it only inside the answer tag, gives \CMOfficialBase{} and \CMOfficialTrained{}, a gain of \CMOfficialGain{}~\CMOfficialCI{}.
    \item[b] The published rubrics with \GDPMaskedCriteria{} criteria judged broken masked and \GDPRejudgedCriteria{} re-graded with a clarified instruction, from an audit of both models' answers by GPT-6.1 Sol and Claude~agents.
  \end{tablenotes}
  \end{threeparttable}
\end{table}

Table~\ref{tab:headline} gives the twelve headline results, and Figure~\ref{fig:headline} draws them.

\paragraph{Coding.} RL-SWE gains \SWEProGain{} points on SWE-bench Pro V2, where base already resolves \SWEProBase\% of tasks, \TBenchGain{} on Terminal-Bench~2.1 and \DeepSWEGain{} on DeepSWE.

\paragraph{Instruction following.} RL-Constraints gains on all four instruction benchmarks: \CCGain{} points on~ComplexConstraints, \AIFGain{} on AdvancedIF, \IFBGain{} on IFBench and \MCGain{} on MultiChallenge. Its training requests were single-turn, yet MultiChallenge, a multi-turn benchmark, gains as much as ComplexConstraints, the benchmark closest to training. IFBench's prompt-level loose accuracy, the metric its paper reports, also gains \IFBPromptLooseGain{}~\IFBPromptLooseCI{}.

\paragraph{Tool use.} RL-ToolUse raises AutomationBench's pass rate from \ABBase\% to \ABTrained\%, and its outcome score, which is zero when any rule of the task is broken, from \ABOutcomeBase{} to \ABOutcomeTrained{} (\ABOutcomeGain{}~\ABOutcomeCI{}). AutomationBench's six departments gain \ABSupportGain{} to \ABHRGain{} points (Appendix~\ref{app:breakdowns}). On HANDBOOK.md, the mean share of each task's criteria that RL-ToolUse meets rises from \HBBase\% to \HBTrained\%.

\paragraph{Charts and documents.} With its Python and crop tools, RL-Charts gains \ChartGain{} points on Chartography and \CMLenientGain{} on ChartMuseum (\CMOfficialGain{} with the official scorer). RL-Docs gains \GDPOfficialGain{} on GDP.pdf with the published rubrics and \GDPCorrectedGain{} with the corrected rubrics; with the published rubrics, it gains \GDPIntentGain{}~\GDPIntentCI{} points on the criteria that require specific content.

\subsection{Other evaluations of the same checkpoints}
\label{sec:other}

\begin{table}[t]
  \centering
  \begin{threeparttable}
  \caption{Every other evaluation of the five checkpoints. ``Setting'' says what differs from the adapter's headline evaluation. Base and trained scores are in percent; gains are in percentage points with 95\% paired task-bootstrap intervals.}
  \label{tab:other}
  \footnotesize
  \setlength{\tabcolsep}{3pt}
  \begin{tabularx}{\linewidth}{@{}l Y r r r@{}}
    \toprule
    Evaluation & Setting & Base & Trained & Gain [95\% CI] \\
    \midrule
    \multicolumn{5}{@{}l}{\textit{RL-SWE}} \\
    SWE-bench Pro V2 & Second run, same settings & \SWEProRunTwoBase & \SWEProRunTwoTrained & \SWEProRunTwoGain{}~\SWEProRunTwoCI \\
    Terminal-Bench 2.1 & Second run, same settings & \TBenchRunTwoBase & \TBenchRunTwoTrained & \TBenchRunTwoGain{}~\TBenchRunTwoCI \\
    Terminal-Bench 2.1 & Another harness, Terminus~2 & \TBenchTerminusBase & \TBenchTerminusTrained & \TBenchTerminusGain{}~\TBenchTerminusCI \\
    SWE-bench Verified & Another benchmark, same harness as SWE-bench Pro V2 & \SWEVerifiedBase & \SWEVerifiedTrained & \SWEVerifiedGain{}~\SWEVerifiedCI \\
    Held-out SWE tasks & Repositories kept out of training; training harnesses and grader & \SWEHeldoutBase & \SWEHeldoutTrained & \SWEHeldoutGain{}~\SWEHeldoutCI \\
    Held-out SWE tasks & As above, at the training effort (medium) & \SWEHeldoutMediumBase & \SWEHeldoutMediumTrained & \SWEHeldoutMediumGain{}~\SWEHeldoutMediumCI \\
    DeepSWE & Five other harness, effort and limit settings; no base run & -- & \DeepSWEFullMini--\DeepSWEVTwo & -- \\
    \multicolumn{5}{@{}l}{\textit{RL-Constraints}} \\
    Complex Constraints dev split & Held-out split of the training data; training judge and token cap & \CCDevBase & \CCDevTrained & \CCDevGainSecond{}~\CCDevGainSecondCI \\
    \multicolumn{5}{@{}l}{\textit{RL-Charts}} \\
    Chartography & Second run, \ChartRerunAttempts{} attempts per task & \ChartRerunBase & \ChartRerunTrained & \ChartRerunGain{}~\ChartRerunCI \\
    Chartography\tnote{a} & No tools & \ChartNoToolsBase & \ChartNoToolsTrained & \ChartNoToolsGain{}~\ChartNoToolsCI \\
    ChartMuseum & No tools & \CMNoToolsLenientBase & \CMNoToolsLenientTrained & \CMNoToolsLenientGain{}~\CMNoToolsLenientCI \\
    CharXiv-RQ & Another benchmark, training tools & \CharXivBase & \CharXivTrained & \CharXivGain{}~\CharXivCI \\
    CharXiv-RQ & Another benchmark, no tools & \CharXivNoToolsBase & \CharXivNoToolsTrained & \CharXivNoToolsGain{}~\CharXivNoToolsCI \\
    ChartQAPro & Another benchmark, no tools & \ChartQAProBase & \ChartQAProTrained & \ChartQAProGain{}~\ChartQAProCI \\
    OCR-Reasoning & Another benchmark, no tools & \OCRReasoningBase & \OCRReasoningTrained & \OCRReasoningGain{}~\OCRReasoningCI \\
    OCRBench v2 (English) & Another benchmark, no tools & \OCRBenchBase & \OCRBenchTrained & \OCRBenchGain{}~\OCRBenchCI \\
    \multicolumn{5}{@{}l}{\textit{RL-Docs}} \\
    GDP.pdf & Temperature \GDPLowTemperature & \GDPTempSixBase & \GDPTempSixTrained & \GDPTempSixGain{}~\GDPTempSixCI \\
    GDP.pdf\tnote{b} & No tools; four context and prompt settings & \GDPNoToolsFirstBase--\GDPNoToolsCheckBase & \GDPNoToolsFirstTrained--\GDPNoToolsCheckTrained & \GDPNoToolsFirstGain{} to \GDPNoToolsPromptGain \\
    AutomationBench & A domain RL-Docs never trained on & \XferPassBase & \XferPassTrained & \XferPassGain{}~\XferPassCI \\
    \bottomrule
  \end{tabularx}
  \begin{tablenotes}[flushleft]\footnotesize
    \item[a] Charts upscaled as in training; at native size the gain is \ChartNoToolsNativeGain{}~\ChartNoToolsNativeCI{}.
    \item[b] All four in the benchmark's no-tools layout: with a smaller budget for the document and the output, \GDPNoToolsFirstGain{}~\GDPNoToolsFirstCI{}; with a larger budget, \GDPNoToolsGain{}~\GDPNoToolsCI{}; adding a training-style system prompt, \GDPNoToolsPromptGain{}~\GDPNoToolsPromptCI{}; adding a self-check turn as well, \GDPNoToolsCheckGain{}~\GDPNoToolsCheckCI{}.
  \end{tablenotes}
  \end{threeparttable}
\end{table}

Table~\ref{tab:other} lists every other evaluation of the five checkpoints, among them Terminal-Bench~2.1 in Terminus~2~\citep{harbor}, SWE-bench Verified~\citep{swebenchverified}, CharXiv-RQ~\citep{charxiv}, ChartQAPro~\citep{chartqapro}, OCR-Reasoning~\citep{ocrreasoning} and OCRBench~v2~\citep{ocrbenchv2}. The three second runs reproduce their headline gains. On data from the training distribution, RL-SWE gains \SWEHeldoutMediumGain{} points on the held-out SWE tasks at the training effort (medium), and RL-Constraints gains \CCDevGainSecond{} on the Complex Constraints dev split at the training token cap. Without its tools, RL-Charts gains \CMNoToolsLenientGain{} points on ChartMuseum. RL-Docs gains \GDPNoToolsFirstGain{} to \GDPNoToolsPromptGain{} points on GDP.pdf without tools and \GDPTempSixGain{} at temperature \GDPLowTemperature{}.

\subsection{The document adapter on AutomationBench}
\label{sec:crossdataset}

RL-Docs was trained on questions about PDF documents, with document tools and no API tools. We ran it on AutomationBench against base on its own deployment, in the same session. It gains \XferPassGain{}~points in pass rate~\XferPassCI{}, \XferOutcomeGain{} in outcome score~\XferOutcomeCI{} and \XferChecksGain{} in the share of required checks met~\XferChecksCI{}. RL-ToolUse, trained on the tool-use data, gains more on each: \ABGain{}, \ABOutcomeGain{} and \ABChecksGain{}~\ABChecksCI{}. The two differ most on the tasks' rules. The share of tasks that break at least one rule rises from \XferRuleBreakBase\% to \XferRuleBreakTrained\% with RL-Docs (\XferRuleBreakGain{}~\XferRuleBreakCI{}) and falls from \ABGuardBase\% to \ABGuardTrained\% with RL-ToolUse (\ABGuardGain{}~\ABGuardCI{}). Section~\ref{sec:behavior} compares their~behavior.
\section{What Changed: A Side-by-Side Analysis}
\label{sec:behavior}

\begin{figure}[t]
  \centering
  \includegraphics[width=\textwidth]{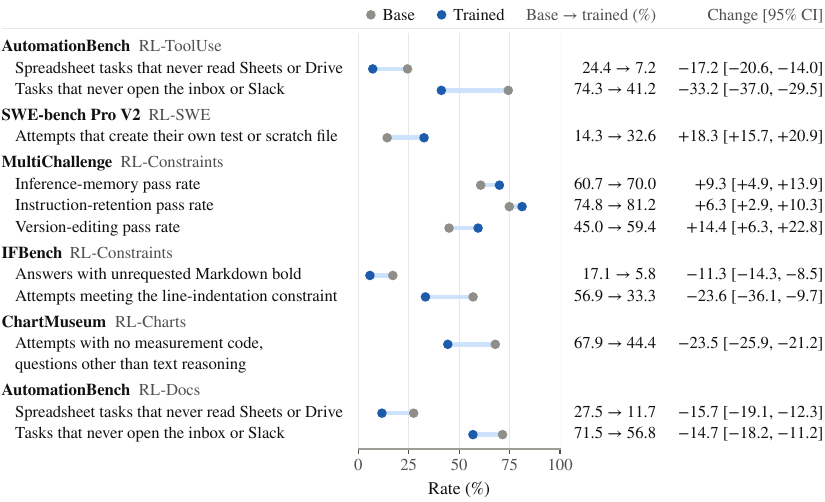}
  \caption{The rates of Table~\ref{tab:modes} over all attempts: base (gray) and the trained adapter (blue), with each paired change in points and its 95\% task-bootstrap interval. Behavior rates are counted by rules on the tool calls, code and answers; the MultiChallenge rows are axis pass rates, and the IFBench line-indentation row is the share of attempts that meet that constraint. Spreadsheet tasks are the AutomationBench tasks seeded with a spreadsheet; RL-Docs is compared with base on its own deployment, and SWE-bench Pro~V2 is measured on its second run (Table~\ref{tab:other}).}
  \label{fig:behavior}
\end{figure}

We paired base and trained attempts on the same tasks and read them side by side: tool calls, code, messages, final answers and, where saved, reasoning. The pairs come from public-benchmark tasks the trained model newly solves: AutomationBench tasks that RL-ToolUse or RL-Docs passed and base failed; DeepSWE, SWE-bench Pro~V2 and Terminal-Bench~2.1 tasks on which every base attempt failed and RL-SWE passed at least once; IFBench and MultiChallenge items that base failed in every attempt and RL-Constraints passed in at least half; and ChartMuseum questions that base failed in both attempts and RL-Charts passed in both. Where a pool held more than \SxsSampleSize{} tasks, we read a random sample of \SxsSampleSize{}; otherwise we read every pair. Claude Opus~5.5 agents described each base failure in a phrase and grouped the phrases into modes (Table~\ref{tab:modes}). Other Claude Opus~5.5 agents, given the first reading's codebook but not each pair's mode, read every pair again and gave the same mode as before to \SxsBlindToolUseSame{} of the \SxsToolUsePairsRead{} RL-ToolUse pairs, \SxsBlindSWESame{} of the \PairSWERead{} RL-SWE pairs and all other pairs; a third agent re-read the data to settle each disagreement. Figure~\ref{fig:behavior} gives the rates of Table~\ref{tab:modes} over all~attempts.

\begin{table}[t]
  \caption{The leading base failure modes in the pairs read (number of pairs; the rest are in the text), the adapter's behavior on the same tasks, the part of the reward that favors it, and the matching rate over all attempts: base\ra{}trained in percent, change in points with its 95\% interval (Figure~\ref{fig:behavior}; Sheets and Drive rates over spreadsheet-seeded tasks; MultiChallenge: axis pass-rate changes).}
  \label{tab:modes}
  \centering
  \scriptsize
  \setlength{\tabcolsep}{3pt}
  \begin{tabularx}{\linewidth}{@{}>{\raggedright\arraybackslash}p{2.75cm} Y Y >{\raggedright\arraybackslash}p{4.5cm}@{}}
    \toprule
    Base failure mode (pairs) & Behavior after RL & Link to training & Rate over all attempts \\
    \midrule
    \multicolumn{4}{@{}l}{\textit{RL-ToolUse on AutomationBench, \SxsToolUsePairsRead{} pairs}} \\
    Data or target not~found~(\SxsToolUseModeNotFound) & Keeps looking: lists the Drive, queries by exact name & Share of requirements met; a stop meets none & Never reads Sheets or Drive: \BehNeverSheetsBase\ra\BehNeverSheetsTrained, \BehNeverSheetsGain{}~\BehNeverSheetsCI \\
    Rule or update kept elsewhere, not read (\SxsToolUseModeMissedRule) & Reads every source first: the whole inbox, every tab & A broken prohibition drops the reward to zero or below & Never opens the inbox or Slack: \BehNeverInboxBase\ra\BehNeverInboxTrained, \BehNeverInboxGain{}~\BehNeverInboxCI \\
    \addlinespace[2pt]
    \multicolumn{4}{@{}l}{\textit{RL-SWE on DeepSWE, SWE-bench Pro V2 and Terminal-Bench 2.1, \PairSWERead{} pairs}} \\
    Checks that missed the error (\PairSWEModeChecks) & Tests the new behavior with a check of its own & Pass only if all hidden tests pass & Own test or scratch file, SWE-bench Pro V2: \PairSWEProOwnFileBase\ra\PairSWEProOwnFileTrained, \PairSWEProOwnFileGain{}~\PairSWEProOwnFileCI \\
    Lost requirement (\PairSWEModeLostReq) & Carries each requirement into the code & One dropped requirement scores zero & -- \\
    \addlinespace[2pt]
    \multicolumn{4}{@{}l}{\textit{RL-Constraints on MultiChallenge and IFBench, \SxsConstraintsPairsRead{} pairs}} \\
    Earlier turn not carried into the answer (\PairMCNewlySolved,~MultiChallenge) & Carries earlier facts, instructions and versions forward & Bonus only if every criterion is met & Axes: inference memory \MCInferenceGain{}~\MCInferenceCI, retention \MCRetentionGain{}~\MCRetentionCI, version editing \MCVersionEditingGain{}~\MCVersionEditingCI \\
    Form rejected by the checker (\PairIFBNewlySolved, IFBench) & Avoids forms a strict reader could count against a rule & The judge reads only the final answer & Unrequested bold: \PairIFBBoldBase\ra\PairIFBBoldTrained, \PairIFBBoldGain{}~\PairIFBBoldCI; line indentation met: \IFBLineIndentBase\ra\IFBLineIndentTrained, \IFBLineIndentGain{}~\IFBLineIndentCI \\
    \addlinespace[2pt]
    \multicolumn{4}{@{}l}{\textit{RL-Charts on ChartMuseum, \SxsCMNewlySolved{} pairs}} \\
    Close call judged by~eye~(\SxsCMModeEye) & Measures the marks in code & All or nothing against an expert answer & No measurement code, questions other than text reasoning: \BehCMNoMeasurementVisualBase\ra\BehCMNoMeasurementVisualTrained, \BehCMNoMeasurementVisualGain{}~\BehCMNoMeasurementVisualCI \\
    Encoding misread (\SxsCMModeEncoding) & Settles what each mark encodes first & As above; a misread encoding scores zero & -- \\
    \addlinespace[2pt]
    \multicolumn{4}{@{}l}{\textit{RL-Docs on AutomationBench, \ExDocsReadPairs{} pairs}} \\
    Source not found (\ExDocsModeSourceCount) & Lists the Drive or queries further, then reads the source & Criteria carry values from the source & Never reads Sheets or Drive: \DocsABNeverSheetBase\ra\DocsABNeverSheetTrained, \DocsABNeverSheetGain{}~\DocsABNeverSheetCI \\
    Inbox not opened (\ExDocsModeMessagesCount) & Reads the inbox before acting & As above & Never opens the inbox or Slack: \DocsABNeverInboxBase\ra\DocsABNeverInboxTrained, \DocsABNeverInboxGain{}~\DocsABNeverInboxCI \\
    One checked detail missed~(\ExDocsModeDetailCount) & Keeps the detail as the task gives it & Each criterion graded on its own & -- \\
    \bottomrule
  \end{tabularx}
\end{table}

\subsection{RL-ToolUse finds the data and the rules before acting}
\label{sec:beh-tooluse}

In \SxsToolUseModeNotFound{} of the \SxsToolUsePairsRead{} pairs, base did not find where the task's data or target lived and stopped or used something of its own; in \SxsToolUseModeMissedRule{} it acted without a rule or update kept in another message, tab or sheet; in \SxsToolUseModeStep{} it did most of the job but dropped or botched one step, and \SxsToolUseModeUnclassified{} pair fit no mode. On spreadsheet-seeded tasks, the share that never reads Google Sheets or Drive falls from \BehNeverSheetsBase\% to \BehNeverSheetsTrained\%; over all tasks, the share that never opens the inbox or Slack falls from \BehNeverInboxBase\% to \BehNeverInboxTrained\%, and RL-ToolUse meets \BehMsgSeedTrained\% of the required checks held only in those messages, against \BehMsgSeedBase\% (\BehMsgSeedGain{}~\BehMsgSeedCI{}).

\textbf{Before.} A finance task asked for statements to clients with outstanding balances and added: ``Follow our standard statement distribution policy.'' The base model read the invoices in Wave, searched the tool catalog for a policy and never opened the inbox, where the accounts-receivable manager's message set a minimum balance below which no statement is sent. It decided to assume standard practice, wrote ``I now have all the data I need.''\ and sent two statements, one for a balance under that minimum, which breaks one of the task's rules.

\textbf{After.} RL-ToolUse listed the inbox, read the policy and sent one statement, for the balance above the minimum.

\textbf{Link to training.} Training rubrics list what must not change as well as what the request requires, and a broken prohibition drops a rollout's reward to zero or below (\S\ref{sec:data}), so a rollout that breaks a rule kept in another service scores below one that read the rule first.

\subsection{RL-SWE checks its work against the requirement}
\label{sec:beh-swe}

In \PairSWEModeLostReq{} of the \PairSWERead{} pairs, base dropped or misread a stated or clearly implied requirement (as in the lost-requirements mode of \citet{crossbenchcoding}). In \PairSWEModeChecks{} it checked its work, but its checks missed the error: they covered only cases it had built, only printed output or only existing tests, or rested on its own assumption (as in that paper's narrow-testing and weak-ground-truth modes). In \PairSWEModeBroke{} it broke behavior that had to keep working, and \PairSWEModeUnclear{} failed for other reasons. On DeepSWE, both models create a test or scratch file of their own in most attempts (\PairDeepSWEOwnFileBase\% and \PairDeepSWEOwnFileTrained\%); on SWE-bench Pro~V2, RL-SWE creates one in \PairSWEProOwnFileTrained\% of attempts against \PairSWEProOwnFileBase\% for base, and runs the tests after its last code change in \BehTestsAfterEditTrained\% against \BehTestsAfterEditBase\% (\BehTestsAfterEditGain{}~\BehTestsAfterEditCI{}).

\textbf{Before.} An Element Web task asked discovery to expose delegated-authentication fields from the stable \texttt{m.authentication} entry, looked up through the SDK constant \texttt{M\_AUTHENTICATION}. The base model read \path{discoveryResult[M_AUTHENTICATION.name]}, ran the existing tests, type checker and linters, and reported the change verified. None of these checks exercised the new field. In this SDK version \texttt{.name} is the unstable key \path{org.matrix.msc2965.authentication}, so the stable entry was never found and both hidden tests that expect it failed.

\textbf{After.} RL-SWE wrote the same lookup, then a temporary test that fed discovery the stable entry. The test failed, and a log line it added showed the unstable key. It read the SDK's \texttt{UnstableValue} class, looked the entry up under both names and deleted the test once it passed.

\textbf{Link to training.} A rollout earns reward only if every hidden test passes, including fail-to-pass tests of behavior the repository's own tests do not cover; one that tests the new behavior can catch such an error before submitting, and the group-relative advantage credits that testing with the fix.

\subsection{RL-Constraints carries earlier turns into the answer}
\label{sec:beh-constraints}

In all \PairMCNewlySolved{} MultiChallenge pairs, base answered the latest turn but dropped an earlier fact, instruction or version, which the benchmark's inference-memory, instruction-retention and version-editing axes test; on the three inference-memory items, \PairMCNoticedBase{} of its \PairMCNoticedAttempts{} attempts mentioned in their reasoning an earlier fact the item turns on, and none passed; the three axes gain \MCInferenceGain{}, \MCRetentionGain{} and \MCVersionEditingGain{} points. In the \PairIFBNewlySolved{} IFBench pairs, base's answers used forms the rule-based checker rejects, such as list numbers or bold markup, and RL-Constraints avoids such forms (unrequested bold \PairIFBBoldBase\%\ra\PairIFBBoldTrained\%). The same caution lowers the share of attempts meeting the line-indentation constraint (each line indented further than the last) from \IFBLineIndentBase\% to \IFBLineIndentTrained\%: in \PairIndentFenceTrained{} of \PairIndentAttempts{} attempts, against \PairIndentFenceBase{} for base, RL-Constraints puts those lines in a code block, whose unindented fences break the pattern.

\textbf{Before.} A MultiChallenge user mentions a hypertension diagnosis and a wish to lose weight in the first turn and later asks for a weekly plan to train with a partner who has asthma and wants to build muscle. Every base attempt plans muscle gain with asthma precautions, and none plans for the user's weight loss; the \PairPlanBaseNoBP{} whose reasoning never returns to the first turn omit blood pressure too.

\textbf{After.} RL-Constraints passes \PairPlanTrainedPasses{} of \MCAttempts{} attempts, exactly those whose reasoning recalls the weight-loss goal, which no base attempt's reasoning does. Its first passing plan warns against holding the breath, which can raise blood pressure, and adds walks for weight loss.

\textbf{Link to training.} Training requests are single-turn, often with pasted source material, and earn a bonus only when every criterion is met (\S\ref{sec:data}), so an answer that drops one requirement loses it. A conversation spreads its requirements over earlier text in the same way.

\subsection{RL-Charts measures the marks and settles what they encode}
\label{sec:beh-charts}

In \SxsCMModeEye{} of the \SxsCMNewlySolved{} pairs, the answer turned on a difference too fine to settle by eye; in \SxsCMModeEncoding{} base read the right marks with the wrong meaning, such as an inverted color key; in \SxsCMModePartial{} it answered from printed labels or outside knowledge instead of the marks, and in \SxsCMModeOther{} it misread the question. Base ran measurement code in \SxsCMPoolBaseMeasured{} of its \SxsCMPoolAttempts{} attempts on these questions and RL-Charts in \SxsCMPoolTrainedMeasured{}. Over all ChartMuseum questions other than text reasoning, the attempts without measurement code fall from \BehCMNoMeasurementVisualBase\% to \BehCMNoMeasurementVisualTrained\%, and in Chartography's second run (Table~\ref{tab:other}) from \BehNoMeasurementBase\% to \BehNoMeasurementTrained\% (\BehNoMeasurementGain{}~\BehNoMeasurementCI{}).

\textbf{Before.} A ChartMuseum scatter plot of OECD countries sets weekly wage against weekly hours and asks which country in a given wage band works the most hours. South Korea's bubble straddles the gridline at the band's lower edge, its center just inside. In both attempts the base model judged the band by eye, from the full chart and one crop that missed South Korea, and answered New Zealand, which works fewer hours; the second attempt named New Zealand and Ireland as tied.

\textbf{After.} RL-Charts found the gridlines in code, detected the bubbles and converted nearby centers to data, putting South Korea inside the band and Israel below it; it answered South Korea both times.

\textbf{Link to training.} Training answers are graded all or nothing against an expert answer that is often an accepted range (\S\ref{sec:data}). On close calls, rollouts that measure earn the reward more often than rollouts that estimate, and the group-relative advantage (\S\ref{sec:recipe}) credits the measuring.

\subsection{RL-Docs finds the data in a domain it never trained on}
\label{sec:beh-docs}

RL-Docs, trained with document tools and no API tools, passes \XferOnlyTrained{} AutomationBench tasks that base fails. In \ExDocsModeSourceCount{} of the \ExDocsReadPairs{} pairs read, base never found the spreadsheet or record that held the data and used values of its own, guessed endpoints or stopped; in \ExDocsModeMessagesCount{} it never opened the inbox that held the procedure or a rule; in \ExDocsModeDetailCount{} it missed one detail a check reads, such as a full name; and in \ExDocsModeStopCount{} it stopped early. On spreadsheet-seeded tasks, the share that never reads Sheets or Drive falls from \DocsABNeverSheetBase\% to \DocsABNeverSheetTrained\%; over all tasks, the share that never opens the inbox or Slack falls from \DocsABNeverInboxBase\% to \DocsABNeverInboxTrained\%.

\textbf{Before.} A finance task asked for invoices to be converted to USD at ``the most recent rates available'' and logged to the accounts-payable tracker. The base model read the invoices and searched for a rate service. Its searches found a Drive endpoint for deleting files but none for listing them, and it concluded that without a listing it could not find the tracker. It used remembered rates and logged wrong amounts to a new spreadsheet; the Drive held the tracker and its dated rates.

\textbf{After.} RL-Docs searched for a Drive listing, found the tracker and reasoned that the rates might be in the spreadsheet; it logged and e-mailed the amounts at the latest dated rates.

\textbf{Link to training.} RL-Docs' criteria carry their own expected values, found by paging, zooming or searching the document (\S\ref{sec:data}); a remembered rate misses such a criterion, so finding the value scores higher. Training had no Drive and no spreadsheet files; looking the value up carried over.

\subsection{A common thread}
\label{sec:thread}

In most pairs, base acted before establishing what the task required: it missed the data or the rule (RL-ToolUse, RL-Docs); checked its work, but not against what was asked (RL-SWE); dropped what an earlier turn set (RL-Constraints); or judged by eye what it could measure (RL-Charts). Every reward grades the final answer or state against the requirements, so rollouts that find, measure or test what a requirement depends on score higher; each adapter takes that step more often. The habit carries over to new tools: RL-Docs, trained without API tools, reads AutomationBench's spreadsheets more often.
\section{Conclusion}
\label{sec:conclusion}

Five LoRA adapters, each trained with RL on one expert-built dataset, raised Nemotron~3.5 Super's headline scores in their domains by \IFBGain{} to \GDPOfficialGain{} points across twelve benchmarks, eight built by other groups and four by Surge AI. The document adapter also gained \XferPassGain{} points on AutomationBench, a domain it never trained on. Side by side, each adapter does more often what base missed, from reading the rules before acting to measuring chart marks in code; on this evidence, the common change is how the model gathers and checks what its tasks need.

\bibliographystyle{plainnat}
{\exhyphenpenalty=50 \interlinepenalty=10000 \raggedright
\bibliography{references}}

\clearpage
\appendix
\section{Training Details}
\label{app:training}

Table~\ref{tab:recipe} lists the training settings that differ between the five runs.

\begin{table}[H]
  \caption{Training settings that differ between the runs. Common to all: a fresh rank-\RecipeLoRARankSWE{} LoRA adapter; CISPO with the importance weight clipped to $[\RecipeClipLow, \RecipeClipHigh]$; the group-mean advantage without standard-deviation scaling; no KL or entropy term; AdamW with $\beta_1=\RecipeAdamBetaOne$, $\beta_2=\RecipeAdamBetaTwo$, $\epsilon=\RecipeAdamEps$, weight decay \RecipeWeightDecay{} and gradient-norm clipping at \RecipeGradClip{}; a learning rate that warms up over the first few updates; training rollouts at temperature \RecipeTrainTemperature{} and top-$p$ \RecipeTrainTopP{}. Soft caps bound only group starts; ``admission held'' pauses new groups while one lags past it; others drop late groups.}
  \label{tab:recipe}
  \centering
  \scriptsize
  \setlength{\tabcolsep}{3pt}
  \begin{tabularx}{\linewidth}{@{}>{\raggedright\arraybackslash}p{2.2cm} Y Y Y Y Y@{}}
    \toprule
    & RL-SWE & RL-Constraints & RL-ToolUse & RL-Charts & RL-Docs \\
    \midrule
    Deployment & early access & documented & early access & documented & documented \\
    Peak learning rate & \RecipeLRSWE & \RecipeLRConstraints & \RecipeLRToolUse & \RecipeLRCharts & \RecipeLRDocs \\
    Schedule to the reported checkpoint & constant & constant & constant to mid-run, then linear decay & constant & constant \\
    Staleness cap (sampler versions) & \RecipeStalenessSWE{} (soft) & \RecipeStalenessConstraints & \RecipeStalenessToolUse{} (soft) & \RecipeStalenessCharts & \RecipeStalenessDocs{} (admission held) \\
    Token cap & \RecipeTurnTokensSWE{} per turn & \RecipeMaxTokensEarlyConstraints{}, then \RecipeMaxTokensConstraints{} per response & \RecipeTurnTokensToolUse{} per turn & \RecipeTurnTokensCharts{} per turn & \RecipeTurnTokensDocs{} per turn \\
    Context & \RecipeContextSWE & one response & \RecipeContextToolUse & \RecipeContextCharts & \RecipeContextDocs \\
    Turn cap & \RecipeTurnCapSWE & one turn & \RecipeTurnCapToolUse & \RecipeTurnCapCharts & \RecipeTurnCapDocs \\
    Episode timeout & \RecipeEpisodeHoursSWE{} h & none & \RecipeEpisodeHoursToolUse{} h & hang guard only & soft notice, then drop \\
    Reasoning effort, training (evaluation) & medium (full) & full (full) & full (full) & full (full) & full (full) \\
    Rollout harness & OpenCode and mini-swe-agent ports & one response & MCP tools in a simulated company & Python kernel and crop tool & document tools \\
    Shaping & none & overlong penalty; a small length term late in the run & long-turn charge; cut-off floor; two over-persistence terms & none & cut-off floor \\
    \bottomrule
  \end{tabularx}
\end{table}

\paragraph{Deployments.} The early-access deployment and the one listed in Tinker's documentation are
\begin{itemize}[leftmargin=*,itemsep=0pt,topsep=2pt]
  \item \texttt{\small\DeployIdEarly} and
  \item \texttt{\small\DeployIdPlain}.
\end{itemize}
They serve different weights: on a fixed \CanaryTokens{}-token sequence, their log-probabilities differ by \CanaryMeanDiff{}~nats per token on average.

\paragraph{Rewards and shaping.} The tool-use reward is the share of required criteria met, less a term for failed content prohibitions; a broken world-state prohibition or an unrequested side effect sets ${R=\RecipeViolationScaleToolUse\,c-\RecipeViolationScaleToolUse}$, and an attempt to reach the environment's backend or to inject instructions sets ${R=\RecipeRewardMinToolUse}$. The constraints run penalizes long failed responses: a response at or below its group's mean reward loses up to \RecipeOverlongAmpConstraints{} of reward once it passes \RecipeOverlongStartConstraints{} tokens, ramping over \RecipeOverlongWidthConstraints{} tokens; late in the run, before the reported checkpoint, a small group-relative length term was added. The tool-use run charges reasoning tokens in a turn beyond \RecipeLongTurnStartToolUse{}, by up to \RecipeLongTurnChargeToolUse{} of advantage over the next \RecipeLongTurnWidthToolUse{} tokens, trains a turn cut off unclosed at ${\min(A,0)-\RecipeCutoffFloorToolUse}$, and weights the outcome of a forced ending by \RecipeForcedScaleToolUse{}. From \RecipeLeversFractionToolUse{} of the way through the run it adds two terms against over-persistence: the turn that ignores the context wrap-up notice trains at ${\min(A,0)-\RecipeWrapupFloorToolUse}$, and tool turns that start beyond \RecipeRunoutStartToolUse{} tokens of context are charged \RecipeRunoutChargeToolUse{}. The document run trains a turn cut at its token cap at ${\min(A,0)-\RecipeCutoffFloorDocs}$.

\section{Evaluation Details and Statistics}
\label{app:eval}

\paragraph{Settings.} Every headline evaluation samples at temperature \EvalTemperature{} and the model's full reasoning effort, with its earlier reasoning kept in context in multi-turn harnesses; attempts are sampled independently. The coding agents get \SWECallCap{} model calls per attempt on SWE-bench Pro V2 and Terminal-Bench~2.1 and \DeepSWECallCap{} on DeepSWE, with \SWETurnTokens{} tokens per turn; Terminal-Bench~2.1 allows \TBenchToolHours{} hours of tool time with a \TBenchBackstopHours{}-hour backstop. AutomationBench allows \ABTurnLimit{} agent steps and HANDBOOK.md \HBTurnCap{} turns and \HBAgentMinutes{} minutes, both with \LongHorizonTurnTokens{} tokens per turn. The chart benchmarks allow \ChartTurnCap{} turns and \ChartContextTokens{} tokens of context, the instruction-following benchmarks \InstructionOutputCap{}~tokens per response, and GDP.pdf \GDPEvalTurnCap{} turns.

\paragraph{Tasks scored.} SWE-bench Pro V2 is scored on \SizeSWEPro{} of its \SizeSWEProTotal{} tasks, because the reference patches of the others fail V2's own tests; DeepSWE on \SizeDeepSWE{} of \SizeDeepSWETotal{}, because one reference solution fails validation; and AdvancedIF on \SizeAIF{} of \SizeAIFTotal{} items, because the judge could not grade two. On SWE-bench Verified, \SWEVerifiedDropped{} of its \SizeSWEVerifiedTotal{} tasks are left out for both models: \SWEVerifiedDroppedInternet{} need the internet at test time, \SWEVerifiedDroppedEmptyPatch{} pass with an empty patch, \SWEVerifiedDroppedGolden{} have a reference patch that fails validation and \SWEVerifiedDroppedInfra{} was lost to an infrastructure failure.

\paragraph{Graders.} The three judged instruction-following benchmarks use Claude Opus~5.5 with each benchmark's own judge prompt; the ComplexConstraints harness grades one criterion per call. The chart benchmarks are graded by GPT-6.1 Sol through their official scoring code, which for ChartMuseum and CharXiv-RQ replaces the official judge model; ChartMuseum's reported scores are format-lenient (Table~\ref{tab:headline}, note~a). HANDBOOK.md is graded by an agent judge on GPT-6.1 Sol; regrading the same trials with the benchmark's release judge, GPT-5.6 Sol, gives a gain of \HBReleaseGain{}~\HBReleaseCI{}. GDP.pdf is graded at high effort with its own per-criterion prompt.

\paragraph{Statistics.} Each interval is a paired task bootstrap: per task, we average each model's attempts and take trained minus base, then resample tasks with replacement \BootResamples{} times with a fixed seed (numpy's default generator, seed~0) and read the 2.5th and 97.5th percentiles. AutomationBench intervals are resampled within its six departments, and the means over AdvancedIF subsets and MultiChallenge axes within subset or axis. Every number is rounded once, half away from zero.

\paragraph{Overlap with the training data.} We compared each headline benchmark and SWE-bench \mbox{Verified} with the training text of the run whose checkpoint it scores: prompts, rubrics and, for the tool-use run, the seeded databases, documents and files of every training task. Text is lower-cased and split into words. Two rules flag an item: the per-instance rule of T\"ulu~3~\citep{tulu3}, under which more than half of the item's tokens are covered by word 8-grams shared with one training instance, and a corpus rule, under which more than half of the item's word 8-grams occur anywhere in the training text. We also count the items that share a word 13-gram with the training text~\citep{gpt3}. As positive controls, \OvlPlantedCopies{} training passages planted into benchmark items all produced 13-gram matches with their source, and all \OvlPlantedLongFlagged{} long passages were flagged. Table~\ref{tab:overlap} gives the results. The corpus rule flags no item, and the per-instance rule flags one ChartMuseum question, which shares a generic question frame with a training prompt about a different chart.

\begin{table}[H]
  \caption{Overlap between each benchmark and the training text of the checkpoint it scores: items sharing any word 13-gram, and the largest share of an item's word 8-grams found in the training text. Gains in the last column are in points, with 95\% intervals.}
  \label{tab:overlap}
  \centering
  \scriptsize
  \setlength{\tabcolsep}{3pt}
  \begin{tabularx}{\linewidth}{@{}>{\raggedright\arraybackslash}p{2.3cm} r r Y@{}}
    \toprule
    Benchmark & Any 13-gram & Max share (\%) & Notes \\
    \midrule
    Terminal-Bench 2.1 & \OvlTBenchAnyThirteen & \OvlTBenchMaxShare & The benchmark's canary string appears in no training prompt \\
    SWE-bench Pro V2 & \OvlSWEProAnyThirteen & \OvlSWEProMaxShare & No SWE training task shares a repository with the three SWE benchmarks \\
    DeepSWE & \OvlDeepSWEAnyThirteen & \OvlDeepSWEMaxShare & \\
    SWE-bench Verified & \OvlSWEVerifiedAnyThirteen & \OvlSWEVerifiedMaxShare & \\
    IFBench & \OvlIFBAnyThirteen & \OvlIFBMaxShare & \\
    AdvancedIF & \OvlAIFAnyThirteen & \OvlAIFMaxShare & Without the 13-gram item: \OvlAIFWithoutThirteenGain{}~\OvlAIFWithoutThirteenCI \\
    MultiChallenge & \OvlMCAnyThirteen & \OvlMCMaxShare & \\
    ComplexConstraints & \OvlCCAnyThirteen & \OvlCCMaxShare & Without the items whose prompt or rubric shares a 13-gram: \OvlCCWithoutThirteenGain{}~\OvlCCWithoutThirteenCI; without the \OvlCCTwins{} items adjudicated as near-duplicates of training prompts: \CCNoTwinsGain{}~\CCNoTwinsCI \\
    AutomationBench & \OvlABAnyThirteen & \OvlABMaxShare & \\
    HANDBOOK.md & \OvlHBThirteenTasks{} tasks & \OvlHBSeedsMaxShare & 13-grams are stock strings, such as field names and timestamps; without those tasks: \OvlHBWithoutThirteenGain{}~\OvlHBWithoutThirteenCI \\
    Chartography & \OvlChartAnyThirteen & \OvlChartMaxShare & Charts matching a benchmark chart were removed before training \\
    ChartMuseum & \OvlCMAnyThirteen & \OvlCMMaxShare & Format-lenient gain without the flagged question, \OvlCMWithoutFlaggedGain{}~\OvlCMWithoutFlaggedCI; without the \OvlCMSharedCharts{} questions whose charts overlap training charts, \CMNoOverlapGain{}~\CMNoOverlapCI \\
    GDP.pdf & none & \OvlGDPCriteriaMaxShare & Prompts and rubrics; training tasks sharing a document were removed before training \\
    \bottomrule
  \end{tabularx}
\end{table}

\paragraph{Tokens.} On the eleven headline benchmarks with per-attempt token records, the trained checkpoints sample \TokCMRatio{} (ChartMuseum) to \TokSWEProRatio{} (SWE-bench Pro V2, second run) times as many tokens per attempt as base, in geometric means over tasks.

\section{Breakdowns}
\label{app:breakdowns}

Tables~\ref{tab:break-a} and~\ref{tab:break-b} split eight of the headline benchmarks by their own labels. Intervals are paired task bootstraps within each subgroup; B/W/T counts the tasks on which the trained model's mean over its attempts is better than, worse than or tied with base's.

\begin{table}[H]
  \caption{Breakdowns of the tool-use, coding and chart headline benchmarks.}
  \label{tab:break-a}
  \centering
  \scriptsize
  \setlength{\tabcolsep}{4pt}
  \begin{tabular}{@{}l r r r r@{}}
    \toprule
    Subgroup & Base & Trained & Gain [95\% CI] & B/W/T \\
    \midrule
    \multicolumn{5}{@{}l}{\textit{AutomationBench (RL-ToolUse), pass rate, by department}} \\
    HR & \ABHRBase & \ABHRTrained & \ABHRGain{}~\ABHRCI & \ABHRBWT \\
    Sales & \ABSalesBase & \ABSalesTrained & \ABSalesGain{}~\ABSalesCI & \ABSalesBWT \\
    Operations & \ABOperationsBase & \ABOperationsTrained & \ABOperationsGain{}~\ABOperationsCI & \ABOperationsBWT \\
    Finance & \ABFinanceBase & \ABFinanceTrained & \ABFinanceGain{}~\ABFinanceCI & \ABFinanceBWT \\
    Marketing & \ABMarketingBase & \ABMarketingTrained & \ABMarketingGain{}~\ABMarketingCI & \ABMarketingBWT \\
    Support & \ABSupportBase & \ABSupportTrained & \ABSupportGain{}~\ABSupportCI & \ABSupportBWT \\
    \multicolumn{5}{@{}l}{\textit{HANDBOOK.md (RL-ToolUse), mean share of criteria met, by business domain}} \\
    HR & \HBHRBase & \HBHRTrained & \HBHRGain{}~\HBHRCI & \HBHRBWT \\
    Logistics & \HBLogisticsBase & \HBLogisticsTrained & \HBLogisticsGain{}~\HBLogisticsCI & \HBLogisticsBWT \\
    Finance & \HBFinanceBase & \HBFinanceTrained & \HBFinanceGain{}~\HBFinanceCI & \HBFinanceBWT \\
    Insurance & \HBInsuranceBase & \HBInsuranceTrained & \HBInsuranceGain{}~\HBInsuranceCI & \HBInsuranceBWT \\
    Medical billing & \HBMedicalBase & \HBMedicalTrained & \HBMedicalGain{}~\HBMedicalCI & \HBMedicalBWT \\
    \multicolumn{5}{@{}l}{\textit{SWE-bench Pro V2 (RL-SWE), resolve rate, by repository language}} \\
    Go & \SWEProGoBase & \SWEProGoTrained & \SWEProGoGain{}~\SWEProGoCI & \SWEProGoBWT \\
    Python & \SWEProPythonBase & \SWEProPythonTrained & \SWEProPythonGain{}~\SWEProPythonCI & \SWEProPythonBWT \\
    TypeScript & \SWEProTypeScriptBase & \SWEProTypeScriptTrained & \SWEProTypeScriptGain{}~\SWEProTypeScriptCI & \SWEProTypeScriptBWT \\
    JavaScript & \SWEProJavaScriptBase & \SWEProJavaScriptTrained & \SWEProJavaScriptGain{}~\SWEProJavaScriptCI & \SWEProJavaScriptBWT \\
    \multicolumn{5}{@{}l}{\textit{Chartography (RL-Charts), accuracy, by professional field}} \\
    STEM & \ChartSTEMBase & \ChartSTEMTrained & \ChartSTEMGain{}~\ChartSTEMCI & \ChartSTEMBWT \\
    Manufacturing and supply chain & \ChartManufacturingBase & \ChartManufacturingTrained & \ChartManufacturingGain{}~\ChartManufacturingCI & \ChartManufacturingBWT \\
    Healthcare & \ChartHealthcareBase & \ChartHealthcareTrained & \ChartHealthcareGain{}~\ChartHealthcareCI & \ChartHealthcareBWT \\
    Finance and investing & \ChartFinanceBase & \ChartFinanceTrained & \ChartFinanceGain{}~\ChartFinanceCI & \ChartFinanceBWT \\
    \bottomrule
  \end{tabular}
\end{table}

\begin{table}[H]
  \caption{Breakdowns of the instruction-following headline benchmarks (RL-Constraints).}
  \label{tab:break-b}
  \centering
  \scriptsize
  \setlength{\tabcolsep}{4pt}
  \begin{tabular}{@{}l r r r r@{}}
    \toprule
    Subgroup & Base & Trained & Gain [95\% CI] & B/W/T \\
    \midrule
    \multicolumn{5}{@{}l}{\textit{MultiChallenge, pass rate per attempt, by axis}} \\
    Inference memory & \MCInferenceBase & \MCInferenceTrained & \MCInferenceGain{}~\MCInferenceCI & \MCInferenceBWT \\
    Instruction retention & \MCRetentionBase & \MCRetentionTrained & \MCRetentionGain{}~\MCRetentionCI & \MCRetentionBWT \\
    Reliable version editing & \MCVersionEditingBase & \MCVersionEditingTrained & \MCVersionEditingGain{}~\MCVersionEditingCI & \MCVersionEditingBWT \\
    Self-coherence & \MCSelfCoherenceBase & \MCSelfCoherenceTrained & \MCSelfCoherenceGain{}~\MCSelfCoherenceCI & \MCSelfCoherenceBWT \\
    \multicolumn{5}{@{}l}{\textit{AdvancedIF, every rubric item met, by subset}} \\
    Carried context & \AIFCarriedBase & \AIFCarriedTrained & \AIFCarriedGain{}~\AIFCarriedCI & \AIFCarriedBWT \\
    Complex instruction following & \AIFComplexBase & \AIFComplexTrained & \AIFComplexGain{}~\AIFComplexCI & \AIFComplexBWT \\
    System steerability & \AIFSteerBase & \AIFSteerTrained & \AIFSteerGain{}~\AIFSteerCI & \AIFSteerBWT \\
    \multicolumn{5}{@{}l}{\textit{ComplexConstraints, every criterion met, by instruction type}} \\
    Implicit requirements & \CCImplicitBase & \CCImplicitTrained & \CCImplicitGain{}~\CCImplicitCI & \CCImplicitBWT \\
    Multistep instructions & \CCMultistepBase & \CCMultistepTrained & \CCMultistepGain{}~\CCMultistepCI & \CCMultistepBWT \\
    Negative instructions & \CCNegativeBase & \CCNegativeTrained & \CCNegativeGain{}~\CCNegativeCI & \CCNegativeBWT \\
    Conditional instructions & \CCConditionalBase & \CCConditionalTrained & \CCConditionalGain{}~\CCConditionalCI & \CCConditionalBWT \\
    \multicolumn{5}{@{}l}{\textit{IFBench, share of constraints met (strict), by constraint family}} \\
    Word-level & \IFBWordsBase & \IFBWordsTrained & \IFBWordsGain{}~\IFBWordsCI & \IFBWordsBWT \\
    Formatting & \IFBFormatBase & \IFBFormatTrained & \IFBFormatGain{}~\IFBFormatCI & \IFBFormatBWT \\
    Counting & \IFBCountBase & \IFBCountTrained & \IFBCountGain{}~\IFBCountCI & \IFBCountBWT \\
    Sentence-level & \IFBSentenceBase & \IFBSentenceTrained & \IFBSentenceGain{}~\IFBSentenceCI & \IFBSentenceBWT \\
    Ratios & \IFBRatioBase & \IFBRatioTrained & \IFBRatioGain{}~\IFBRatioCI & \IFBRatioBWT \\
    Repeating & \IFBRepeatBase & \IFBRepeatTrained & \IFBRepeatGain{}~\IFBRepeatCI & \IFBRepeatBWT \\
    One-off tasks & \IFBCustomBase & \IFBCustomTrained & \IFBCustomGain{}~\IFBCustomCI & \IFBCustomBWT \\
    \bottomrule
  \end{tabular}
\end{table}

\end{document}